\pdfoutput=1

\documentclass[11pt]{article}

\usepackage{acl}

\usepackage{times}
\usepackage{latexsym}
\usepackage{subcaption}
\usepackage{amsmath}
\usepackage{booktabs}
\usepackage{longtable}
\usepackage{enumitem}
\usepackage{tabularray}
\usepackage{array}
\UseTblrLibrary{booktabs}
\SetTblrInner{rowsep=2pt,colsep=4pt}

\usepackage[T1]{fontenc}
\usepackage[utf8]{inputenc}
\usepackage{inconsolata}
\usepackage{graphicx}

\title{When Linguistic and Internal Confidence Diverge in Large Language Models}

\author{
  \textbf{Hefan Zhang\textsuperscript{1}},
  \textbf{Bingquan Zhang\textsuperscript{1}},
  \textbf{Ming Cheng\textsuperscript{1}},
  \textbf{Saeed Hassanpour\textsuperscript{1}},
\\
  \textbf{Weicheng Ma\textsuperscript{2}},
  \textbf{Soroush Vosoughi\textsuperscript{1}}
\\
  \textsuperscript{1}Dartmouth College,
  \textsuperscript{2}Oakland University
\\
  \small{
    \href{mailto:hefan.zhang.gr@dartmouth.edu}{\texttt{\{hefan.zhang.gr,}}
    \href{mailto:soroush.vosoughi@dartmouth.edu}{\texttt{soroush.vosoughi\}}}
    \texttt{@dartmouth.edu}
  }
}

\begin{document}
\maketitle

\begin{abstract}
Users often ask large language models (LLMs) to report how confident they are, but it is unclear whether such \emph{linguistic} confidence tracks the model's \emph{internal} confidence. We study this question across 8 classification tasks, 2 generation tasks and 30 models from three families. For classification, we compare linguistic confidence with logits-based confidence along three axes: association, magnitude agreement and calibration. For generation, we test whether linguistic confidence tracks semantic-entropy-based uncertainty. The axes frequently diverge. Instance-level association is weak on average, although it improves on easier items and for stronger base models. Instruction-tuned models often report higher confidence and sometimes show higher association, but they also have larger confidence gaps and worse calibration. Prompt design mostly changes the distribution of reported confidence. Attitude cues inflate confidence without improving alignment, while score exemplars can preserve rank-order signal when they avoid collapsed confidence values. Regression analyses show that distributional properties of confidence scores explain much of the observed alignment pattern, with model metadata playing a smaller role after controls. These results support a lossy-channel view of linguistic confidence. A more dispersed verbal confidence distribution can carry useful rank information, but it does not make the scores calibrated. Linguistic confidence should therefore be evaluated with multi-axis diagnostics before being used in downstream reliability pipelines.
\end{abstract}

\section{Introduction}

Large language models (LLMs) are increasingly used not only for their answers, but also for confidence signals that help decide when an answer should be trusted, filtered, abstained from, or combined with other samples \cite{geng2023survey, liu2025uncertainty}. Internal signals such as token-level probabilities can support confidence-weighted voting, selective prediction and test-time selection \cite{fu2025deep}. In many deployed systems, however, these internal signals are not available. Users and practitioners are left with linguistic confidence, the model's self-reported confidence in language or as a numeric score.

Prior work has shown that verbalized confidence is fragile. It can be sensitive to prompting and often becomes overconfident after instruction tuning \cite{mielke2022reducing, xiong2023can, groot2024overconfidence, xu-etal-2025-language}. Much of this work evaluates confidence as an estimator of correctness. This is important, but it leaves a separate question open. When a model says it is confident, is that statement aligned with the model's own internal probability signal, or is it a separate user-facing behavior?

We study this cross-channel question. Our goal is not to treat logits-based confidence as ground-truth uncertainty. Softmax probabilities are themselves imperfect and often miscalibrated. Instead, we use them as an internally accessible confidence channel and ask when linguistic confidence agrees with that channel. For generation tasks, where answer-token probabilities are not always well defined, we use semantic entropy as the internal uncertainty proxy \cite{kuhn2023semantic}.

We evaluate linguistic confidence along three axes. Association measures whether linguistic and internal confidence move together across instances. Magnitude agreement measures whether the two scores are close on a common scale. Calibration measures whether reported confidence agrees with empirical correctness. These axes are not interchangeable. A model can have high association while still being overconfident, or it can have small average distance while carrying little instance-level information.

Across 10 tasks and 30 models, we find that instance-level association is weak on average. At the same time, the weakness is structured rather than random. Easier tasks and stronger base models tend to show higher association. Instruction-tuned models often report higher confidence, but their apparent gains in association do not reliably produce better calibration. Prompt changes also have limited leverage. They can shift the mean and variance of linguistic confidence, but attitude cues that raise verbal confidence do not make it more grounded in internal probabilities.

The main empirical pattern is distributional. When linguistic confidence collapses into a narrow range, often near high values, it cannot express instance-level differences. When it is more dispersed, it can preserve rank-order information and improve association. This does not mean that more dispersion fixes calibration. The same dispersion that helps correlation can also increase magnitude mismatch if the scores are shifted or nonlinear. We therefore interpret linguistic confidence as a lossy channel from internal confidence to user-facing text. The channel can carry useful rank signal in some regimes, but it should not be read as a calibrated probability without validation.

This paper makes three contributions. We provide a systematic cross-channel evaluation of linguistic and internal confidence across model families, tasks and prompting regimes. We show that association, magnitude agreement and calibration diverge in important cases, especially for instruction-tuned models. We identify distributional properties of confidence scores as dominant observable drivers of alignment and translate this finding into practical guidance for reliability pipelines. Code is available at \href{https://github.com/HF-heaven/Correlation-between-Confidence-Measurements}{https://github.com/HF-heaven/Correlation-between-Confidence-Measurements}.

\section{Related Work}
\label{sec:related_work}

\paragraph{Uncertainty estimation in LLMs.}
Confidence and uncertainty estimation are central to reliable LLM deployment \cite{huang2024survey, liu2025uncertainty}. Existing methods can be grouped into logits-based, linguistic and consistency-based approaches \cite{geng2023survey}. Logits-based methods use white-box access to output distributions, including softmax probabilities, entropy or logit margins \cite{duan2023shifting, kuhn2023semantic, huang2023look}. Linguistic methods ask the model to express confidence directly in words or scores \cite{mielke2022reducing, xiong2023can}. Consistency-based methods estimate uncertainty from output stability under sampling or perturbation \cite{manakul2023selfcheckgpt, lin2023generating}. These signals differ in access requirements and in what part of the model behavior they measure.

\paragraph{Verbalized and internal confidence.}
Prior work studies both the calibration of verbalized confidence and its relationship to internal uncertainty. \citet{xiong2023can, groot2024overconfidence} show that linguistic confidence can be overconfident and prompt-sensitive. Some other studies connect language-model confidence to human confidence or propose calibration interventions for verbalized scores \cite{mielke2022reducing, xu-etal-2025-language}. These studies usually judge confidence by its relationship to correctness, often through ECE, AUROC, or related calibration metrics. \citet{lin2022teaching} examine whether models can express calibrated uncertainty in words, while \citet{tian2023just} compare verbalized scores with conditional probabilities in models fine-tuned with human feedback. \citet{kumar2024confidence} directly study confidence-probability alignment using elicited certainty scores and token probabilities. \citet{yona2024faithfully} evaluate whether linguistic hedging faithfully reflects intrinsic uncertainty. Most closely related to our generation setting, \citet{ji2025calibrating} analyze mismatch between verbal and semantic uncertainty and use it to predict hallucinations. More broadly, the sensitivity of LLM behavior to linguistic framing \cite{tian-etal-2026-lost} raises questions about whether expressed confidence remains aligned with internal confidence across settings.

Our work extends this literature through a systematic multi-axis analysis across 30 models, 10 tasks, classification and generation settings, and multiple prompting regimes. Rather than treating cross-channel correlation as calibration, we report two separate measurements: association describes agreement between linguistic confidence and the selected internal proxy, whereas calibration evaluates confidence against correctness. Neither measurement enables the other. We additionally examine how task difficulty, model properties, prompting, and confidence-score distributions are associated with these outcomes.

\paragraph{Failure modes and downstream use.}
Cross-channel differences are described as distributional shift, magnitude disagreement, or cross-channel misalignment rather than miscalibration. We reserve \emph{calibration} for comparisons against correctness. A linguistic score may collapse into a narrow range, differ systematically in magnitude from an internal proxy, or fail to preserve instance ordering; these cases motivate different deployment checks. For white-box models, internal confidence and related uncertainty signals
support selection, confidence-weighted voting, abstention, reranking,
selective prediction, and curriculum-based post-training data ordering
\cite{brown2024large,cobbe2021trainingverifierssolvemath,
kamath2020selective,cole2023selectively,fu2025deep,
jia-etal-2026-makes}. The reliability of verifier-based selection is likewise sensitive to problem difficulty and model capability \cite{zhou2026variation}. For black-box systems, linguistic confidence should be validated on held-out examples before it controls downstream decisions.

\section{Confidence Measurements and Evaluation}
\label{sec:measurement}

We compare linguistic confidence with internal confidence proxies. For classification tasks, the internal proxy is logits-based confidence over the candidate answers. For generation tasks, the internal proxy is semantic entropy over sampled responses.

\subsection{Linguistic Confidence}
\label{sec:linguistic_confidence}

We obtain linguistic confidence by asking the model to answer the task and report a confidence score for its own answer \cite{groot2024overconfidence, xiong2023can}. The prompt defines 0 as no belief in the answer and 10 as full confidence. We extract the generated score and divide it by 10 when comparing it with probabilities. Prompt templates are described in Appendix~\ref{appendix:prompt}.

\subsection{Internal Confidence Proxies}
\label{sec:internal_confidence}

For classification tasks, output formats constrain the answer to a fixed candidate set. We extract logits for the option tokens and compute a softmax over the candidate set. We use the resulting probability of the selected answer as logits-based confidence.

This quantity should not be interpreted as ground-truth uncertainty. Softmax probabilities can be miscalibrated, and they do not capture every form of epistemic uncertainty. In this work, they serve as an internal confidence proxy that can be measured in white-box models. The study is therefore about fidelity between two available channels, internal probability signals and linguistic self-reports.

For generation tasks, the key answer is not always tied to one token. We therefore use semantic entropy \cite{kuhn2023semantic}. Given a prompt $x$, we sample $M$ responses and cluster semantically equivalent outputs using an NLI model. The probability of a semantic cluster is estimated by summing the sequence probabilities of its members,
\begin{equation}
    p(c \mid x) = \sum_{s \in c} \prod_i p(s_i \mid s_{<i}, x).
\end{equation}
Semantic entropy is then computed over the clusters,
\begin{equation}
    SE(x) \approx -|C|^{-1} \sum_{i=1}^{|C|} \log p(C_i \mid x),
\end{equation}
where $C$ is the set of semantic equivalence classes. Higher entropy means lower confidence, so we use negated semantic entropy when computing association with linguistic confidence.

\subsection{Evaluation Axes}
\label{sec:evaluation_axes}

We evaluate confidence alignment along three axes. Association is measured with Pearson correlation and Spearman correlation. It captures whether two confidence signals vary in the same direction across instances. Magnitude agreement is measured by Euclidean distance after mapping linguistic confidence to $[0,1]$. It captures whether the two scores are close in value. Calibration is measured with expected calibration error (ECE) \cite{guo2017calibration}. It captures whether reported confidence matches empirical accuracy.

The three-axis analysis applies directly to classification tasks, where logits-based confidence and empirical accuracy are defined on the same prediction. For generation tasks, semantic entropy is not on the same probability scale as linguistic confidence and correctness is less uniform across free-form responses. We therefore use association with negated semantic entropy as the primary cross-channel metric for generation, and use distance and ECE for the classification setting.

These axes answer different questions. Correlation can be high even when linguistic confidence is systematically too large. Distance can be small when both scores are concentrated, even if neither carries much instance-level information. ECE measures agreement with correctness rather than agreement between confidence channels. We report these metrics jointly to avoid treating one form of agreement as full reliability.

\section{Experimental Setting}
\label{sec:experimental_setting}

We evaluate 8 classification tasks and 2 generation tasks. The classification benchmarks cover syntactic judgment (\textit{CoLA}), natural language inference (\textit{QNLI}), paraphrasing (\textit{QQP}) \cite{wang2018glue}, causal reasoning (\textit{Cause And Effect}), lexical reasoning (\textit{Conceptual Combinations} and \textit{Ruin Names}), temporal reasoning (\textit{Temporal Sequences}) \cite{suzgun2022challenging}, and broad knowledge (\textit{MMLU}) \cite{hendrycks2021measuringmassivemultitasklanguage}. The generation tasks are \textit{CoQA} \cite{reddy2019coqa} and \textit{TriviaQA} \cite{joshi2017triviaqa}. For computational tractability, only the five evaluation sets containing more than 200 examples (CoLA, QNLI, QQP, MMLU, and Temporal Sequences) are randomly downsampled to 200 instances. All other tasks use their complete evaluation sets. Repeated-subsampling results and reproducibility details are reported in Appendix~\ref{app:dataset_stability}.

We use 30 publicly available models from the LLaMA, Mistral and Qwen families \cite{touvron2023llama,touvron2023llama2,grattafiori2024llama,jiang2023mistral7b,jiang2024mixtral,qwen2025qwen25technicalreport,yang2025qwen3technicalreport}. The models cover different sizes, versions and tuning types. Table~\ref{tab:model_seletion} lists all models.

We use standardized prompt templates for classification and generation. For base models, we use two-shot examples to constrain the answer and confidence format. For instruction-tuned models, we place the output format requirement in the system prompt. We also test prompt robustness through confidence-exemplar search, confidence-elicitation ablation and attitude-based perturbations. The details are in Appendix~\ref{appendix:prompt} and Appendix~\ref{appendix:effect_of_prompt}.

\section{Results}
We organize the results from coarse to fine. We begin with a pooled descriptive overview, then examine instance-level association and decompose the results by task, model, generation setting, and prompt design. We subsequently evaluate these factors jointly using descriptive regression models. This organization distinguishes aggregate patterns from the stratified evidence relevant to instance-level decisions.
\subsection{Overall Alignment: Confidence Measures Align Weakly}
\label{Chapter:overall_correlation}

\begin{figure}[t]
  \includegraphics[width=\linewidth]{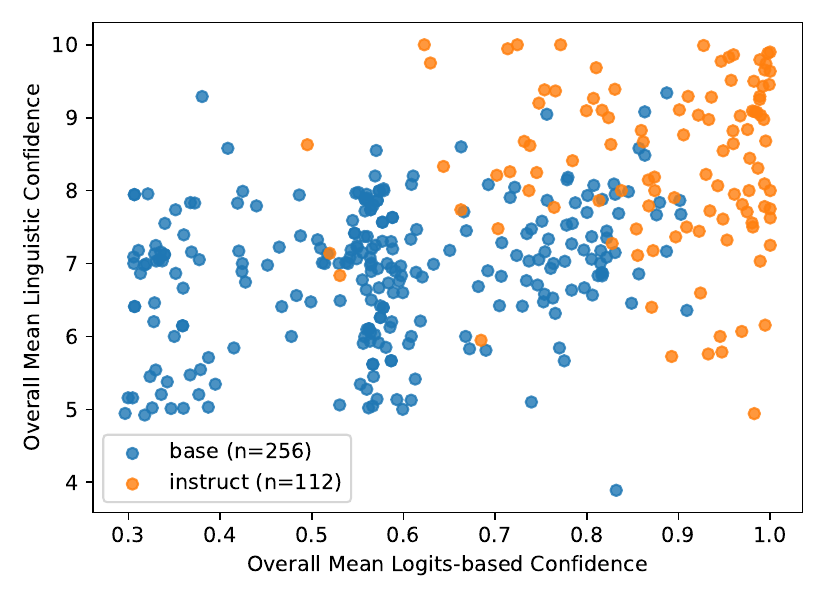}
  \caption{Mean logits-based confidence and mean linguistic confidence across task, model and prompt settings. Each point is one setting. Instruction-tuned models tend to have higher values on both axes. The overall association is statistically significant, but the split by model type shows that base models have weak positive association while instruction-tuned models have almost no aggregate association.}
  \label{fig:confidence_scatter}
\end{figure}

We first average logits-based confidence and linguistic confidence for each task, model and prompt setting. Figure~\ref{fig:confidence_scatter} shows that instruction-tuned models generally produce higher values on both axes than base models. The overall Pearson correlation is $r = 0.4830$ with $p = 1.14 \times 10^{-22}$. The corresponding Spearman correlation is $\rho = 0.4711$ with $p = 1.65 \times 10^{-21}$.

This aggregate trend hides an important split. Base models show a weak but significant positive association, with $r_b = 0.2607$ and $p = 2.40 \times 10^{-5}$, and $\rho_b = 0.2161$ with $p = 4.98 \times 10^{-4}$. Instruction-tuned models show negligible association, with $r_i = -0.0048$ and $p = 0.961$, and $\rho_i = 0.0422$ with $p = 0.665$.

The split suggests that instruction tuning changes how confidence is expressed. Instruction-tuned models often have logits-based probabilities concentrated near 1 and linguistic confidence concentrated near high values. This compression reduces observable association because many distinct examples receive similar confidence values. The result is not that base models are reliable. Rather, base models retain more variation in both channels, which allows weak association to remain visible.

At the instance level, association is weaker than the aggregate plot suggests. Across settings, the average Pearson correlation is $r = 0.135$. This gap between aggregate and instance-level behavior is central to our findings. Linguistic and internal confidence can move together in broad averages while diverging on individual examples, which is the level needed for abstention, filtering and selection.

\subsection{Task-Level Variation: Association Tends to Be Higher on Easier Tasks}
\label{sec:task_difficulty}

\begin{figure}[t]
  \includegraphics[width=\linewidth]{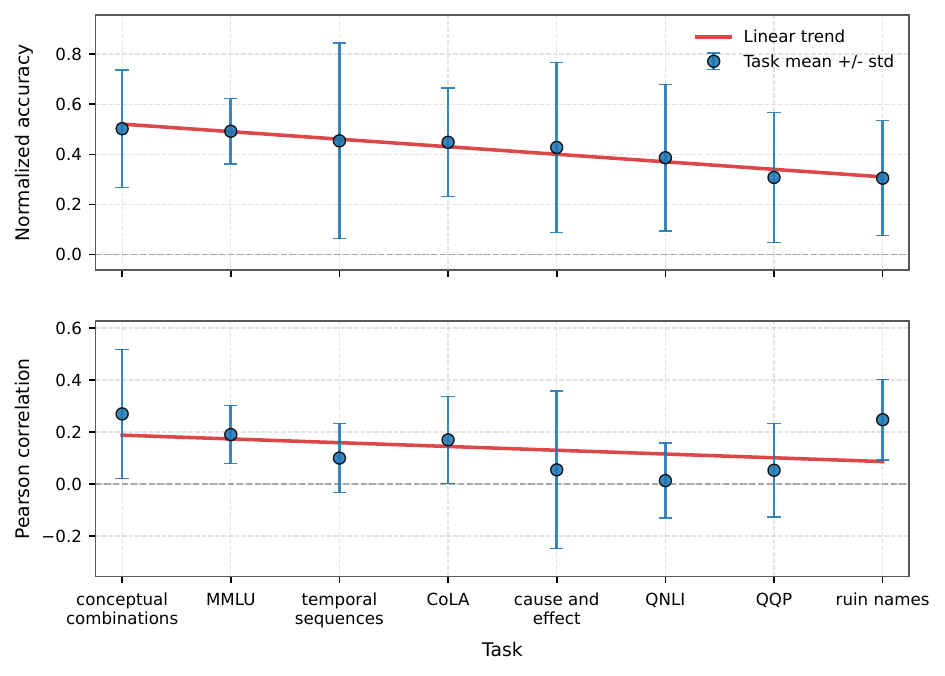}
  \caption{Task difficulty and confidence association. Tasks are ordered by decreasing normalized accuracy in the top panel. The bottom panel shows the Pearson correlation between linguistic confidence and logits-based confidence. Markers show per-task means and error bars show $\pm 1$ standard deviation across models. Association tends to weaken as task difficulty increases.}
  \label{fig:corrbyperformance}
\end{figure}

We next examine whether task difficulty affects the relationship between linguistic and logits-based confidence. Since tasks have different numbers of answer choices, we use chance-corrected normalized accuracy:
\begin{equation}
\mathrm{NormAcc}=\frac{\mathrm{Acc}-1/K}{1-1/K},
\label{eq:norm_acc}
\end{equation}
where $K$ is the number of answer options. This maps random guessing to 0 and perfect accuracy to 1, making binary and four-choice tasks more comparable.

Across all eight tasks, the pooled task-level relationship between normalized accuracy and mean confidence association is positive but not statistically reliable (slope $=0.360$, 95\% CI $[-0.842,1.562]$, $p=0.491$). Excluding \textit{Ruin Names}, the estimated slope is larger (slope $=1.067$, 95\% CI $[0.075,2.058]$, $p=0.0396$). We report this exclusion as a sensitivity analysis rather than as the primary result.

We also estimate the relationship separately within each model--prompt setting. Thirty of 45 slopes are positive (66.7\%; sign-test $p=0.0179$), but the mean slope is small and uncertain (mean $=0.034$, 95\% CI $[-0.155,0.223]$). After averaging prompts within models, 20 of 29 model-level slopes are positive (69.0\%; sign-test $p=0.0307$; mean $=0.121$, 95\% CI $[-0.104,0.347]$). The direction therefore appears in a majority of stratified analyses, but its magnitude is heterogeneous across tasks and models.

\textit{Ruin Names} is an outlier. It has relatively low accuracy but high association between linguistic and logits-based confidence. We avoid interpreting this as good calibration. The task is built around humorous edits of names, so confidence may track surface familiarity or orthographic similarity even when the model chooses the wrong option. This shows why association must be read together with accuracy and calibration.

\subsection{Model-Level Variation: Model Properties Affect Alignment}
\label{Chapter:class_corr_model}

\begin{figure}[t]
  \centering
  \begin{subfigure}[t]{0.235\textwidth}
    \centering
    \includegraphics[width=\linewidth]{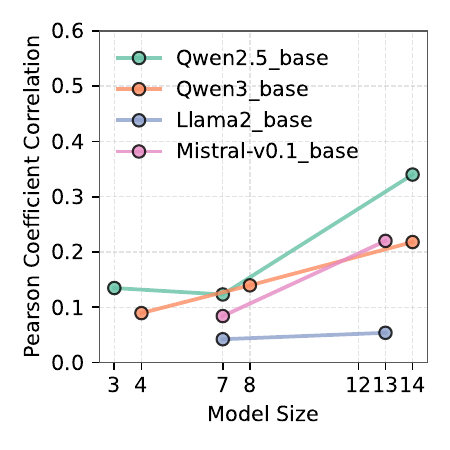}
    \caption{Base models}
    \label{fig:left_wothoutbar}
  \end{subfigure}
  \begin{subfigure}[t]{0.235\textwidth}
    \centering
    \includegraphics[width=\linewidth]{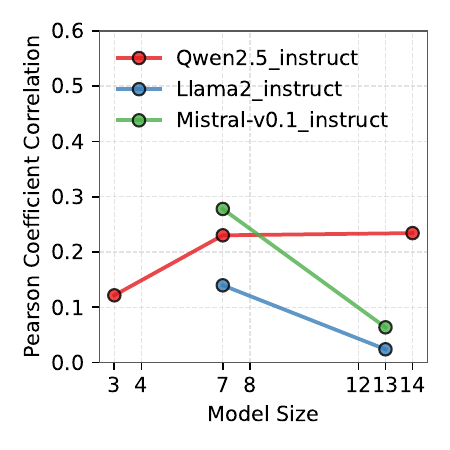}
    \caption{Instruction-tuned models}
    \label{fig:right_wothoutbar}
  \end{subfigure}
  \caption{Model size and association between linguistic and logits-based confidence. Association tends to rise with model size for base models, but it plateaus or decreases for instruction-tuned models.}
  \label{fig:corrbymodel}
\end{figure}

We compare confidence association across model configurations. Figure~\ref{fig:corrbymodel} shows different scaling trends for base and instruction-tuned models. For base models, larger models tend to have higher association. For instruction-tuned models, association is flat or decreases.

\begin{figure}[t]
  \centering

  \begin{subfigure}[t]{0.48\columnwidth}
    \centering
    \includegraphics[width=\linewidth]
    {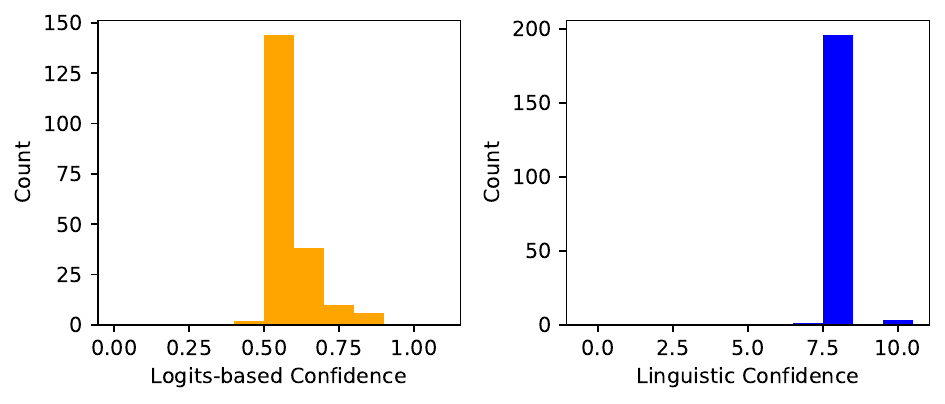}
    \caption{Mistral-7B (base)}
    \label{fig:mistral7b_base_cola}
  \end{subfigure}
  \hfill
  \begin{subfigure}[t]{0.48\columnwidth}
    \centering
    \includegraphics[width=\linewidth]
    {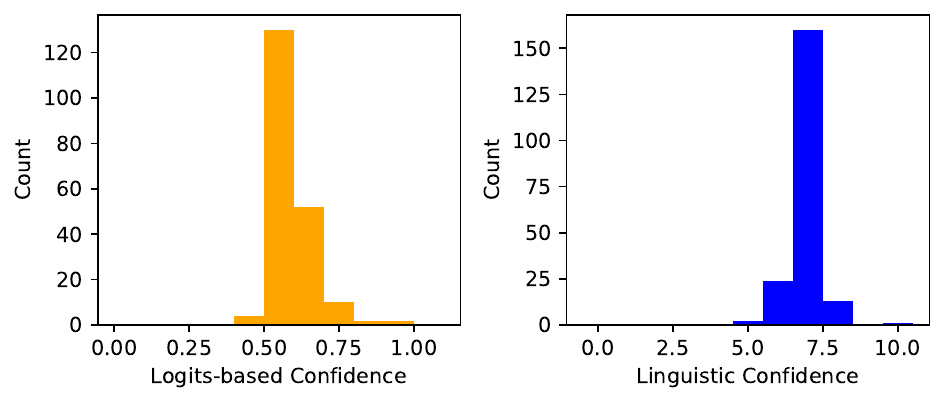}
    \caption{Mixtral-8x7B (base)}
    \label{fig:mixtral_base_cola}
  \end{subfigure}

  \par\vspace{1mm}

  \begin{subfigure}[t]{0.48\columnwidth}
    \centering
    \includegraphics[width=\linewidth]
    {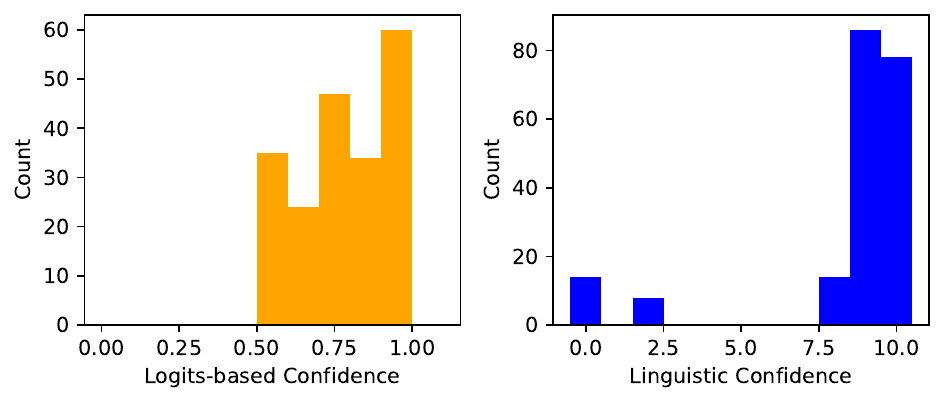}
    \caption{Mistral-7B (instruct)}
    \label{fig:mistral7b_instruct_cola}
  \end{subfigure}
  \hfill
  \begin{subfigure}[t]{0.48\columnwidth}
    \centering
    \includegraphics[width=\linewidth]
    {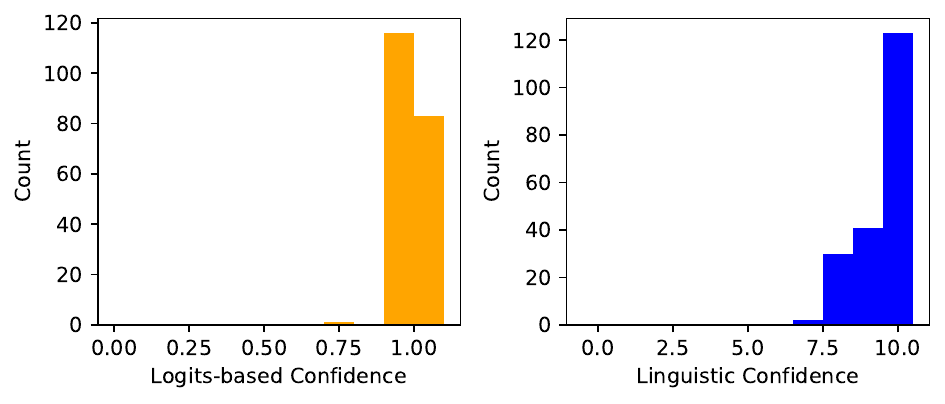}
    \caption{Mixtral-8x7B (instruct)}
    \label{fig:mixtral_instruct_cola}
  \end{subfigure}

  \caption{
    Representative distributions of logits-based and linguistic confidence
    on CoLA for Mistral-v0.1 models. The top row shows base models under
    Base Prompt~1, and the bottom row shows instruction-tuned models.
  }
  \label{fig:mistral_cola_distributions}
\end{figure}

A distributional explanation fits this pattern. Smaller base models often output linguistic confidence concentrated around 8 or 10, which leaves little instance-level signal. Larger base models produce more varied linguistic confidence, which can preserve more rank-order information. In contrast, larger instruction-tuned models often have logits-based confidence values near 1. This saturation reduces internal variation and weakens correlation with linguistic confidence. Figure~\ref{fig:mistral_cola_distributions} provides a representative comparison between base and instruction-tuned Mistral-family models, while additional model-family examples are provided in Appendix~\ref{appendix:logits_conf_distribution}.

\begin{table}[t]
\centering
\small
\setlength{\tabcolsep}{4pt}
\begin{tabular}{lrrrr}
\toprule
Metric & mean$_\text{instruct}$ & mean$_\text{base}$ & $t$ & $p$ \\
\midrule
$r$ & 0.172 & 0.118 & 1.95 & 5.4e-02 \\
Distance & 0.271 & 0.241 & 2.03 & \underline{4.4e-02} \\
ECE(linguistic) & 0.287 & 0.191 & 6.02 & \underline{2.5e-08} \\
\bottomrule
\end{tabular}
\caption{Paired t-test results comparing instruction-tuned and base models. Underlined values indicate significant differences with $p<0.05$.}
\label{tab:instruct_base_ttest}
\vspace{-2mm}
\end{table}

Table~\ref{tab:instruct_base_ttest} shows the same point from another angle. Instruction-tuned models have slightly higher Pearson correlation than base models, but the difference is not significant at the 0.05 level. Their distance and linguistic ECE are significantly larger. Thus, instruction tuning can make the confidence channels appear more associated in some settings while making the reported scores less calibrated.

Across families, we observe a general increase in association from LLaMA to Mistral and Qwen, together with lower distance and ECE. Within-family version updates are less consistent. LLaMA-3.1 improves association over LLaMA-3, while Qwen3 shows lower association and larger distance than Qwen2.5. The appendix analysis (Appendix~\ref{appendix:ttest_family_version}) suggests that these changes are closely tied to shifts in confidence dispersion rather than family or version alone.

\subsection{Generation Setting: Association Remains Task- and Model-Dependent}
\label{sec:generation_tasks}

For generation tasks, we compare linguistic confidence with negated semantic entropy. We sample five responses per prompt, cluster semantically equivalent responses with \texttt{DeBERTa-xlarge-MNLI} \cite{he2021deberta}, compute semantic entropy and average the linguistic confidence across the sampled responses. As noted in Section~\ref{sec:evaluation_axes}, these results are evaluated through association because semantic entropy is an uncertainty score rather than a calibrated probability on the linguistic-confidence scale.

The full results are in Appendix~\ref{appendix:gen_results}. The main pattern is that instruction-tuned models usually show stronger association than their base counterparts on generation tasks. TriviaQA also shows higher association than CoQA. This likely reflects the difference between factoid questions with clearer answers and conversational questions with more context-dependent answer forms. These generation results support the same general conclusion as the classification results. Linguistic confidence can carry some rank-order information, but this does not establish calibrated uncertainty.

\subsection{Prompt-Level Variation: Prompt Design Mainly Changes Score Distributions}
\label{sec:prompt_effects}

We test three prompt-related factors. First, we search over few-shot confidence exemplars. Second, we remove confidence elicitation to test whether asking for confidence perturbs logits. Third, we prepend attitude cues such as criticism, approval and irrelevant text.

The exemplar search shows that confidence examples affect both task performance and association. Adjacent but non-identical scores work better than identical scores. Identical exemplars tend to collapse the produced confidence distribution, which weakens instance-level association. Very large exemplar scores push outputs toward high confidence values and also reduce useful variation. This supports the distributional view that linguistic confidence needs enough spread to preserve rank information.

The ablation study shows that confidence elicitation does not meaningfully perturb the internal probability signals. Removing the confidence query does not change accuracy, and it has limited effects on logits statistics. For instruction-tuned models, no tested metric changes significantly.

The attitude perturbation results are more concerning. Criticism, approval and irrelevant cues all raise reported linguistic confidence, but they reduce association with logits-based confidence. A calibration instruction asking the model to reflect final-layer probabilities does not significantly improve the relationship. Prompt tone therefore changes how confident the model sounds without making the confidence more grounded in internal probabilities. Full results are in Appendix~\ref{appendix:effect_of_prompt}.

\section{Distributional Properties Explain Much of the Pattern}
\label{sec:distributional_properties}

\subsection{Regression Modeling}
\label{sec:regression}

We fit regression models to identify which variables predict confidence alignment. We model three outcomes. The first is the correlation between linguistic and logits-based confidence, where higher is better. The second is the distance between the two scores, where lower is better. The third is a confidence-to-confidence ECE that replaces accuracy with logits-derived confidence, where lower is better. For each outcome, we fit ordinary least squares and ridge regression with standardized features and five-fold cross-validation. OLS coefficients are descriptive because one-hot encodings and correlated features introduce collinearity. Ridge coefficients are used for stable effect ranking.

\begin{table}[t]
\centering
\small
\begin{tabular}{lcccc}
\toprule
Outcome & $n$ & Predictors & $R^2$ & Adj. $R^2$ \\
\midrule
\textsc{Dist} & 363 & 28 & 0.570 & 0.536 \\
\textsc{ECE} & 363 & 28 & 0.644 & 0.615 \\
\textsc{Correlation} & 363 & 28 & 0.498 & 0.459 \\
\bottomrule
\end{tabular}
\caption{OLS fit statistics for distance, confidence-to-confidence ECE and correlation between linguistic and logits-based confidence.}
\label{tab:fit-specified}
\end{table}

Table~\ref{tab:fit-specified} shows that the predictors jointly account for a substantial fraction of the variation in all three outcomes. The OLS models attain $R^2$ values of 0.570 for distance, 0.644 for ECE-style cross-channel mismatch, and 0.498 for correlation.

The main result is that distributional statistics dominate. Larger linguistic dispersion is associated with higher correlation (standardized $\beta=0.194$, 95\% CI for the unstandardized coefficient $[0.012,0.070]$, $p=0.0058$), indicating that greater score variability can help preserve rank-order signal. The same dispersion is also associated with greater distance (standardized $\beta=0.780$, $p<0.001$) and higher confidence-to-confidence ECE (standardized $\beta=0.318$, $p<0.001$). Dispersion is therefore not a universal fix: it can improve association while worsening magnitude agreement and cross-channel ECE. Linguistic mean has a similar role. Higher mean confidence is associated with greater distance (standardized $\beta=0.228$, $p<0.001$) and higher confidence-to-confidence ECE (standardized $\beta=0.395$, $p<0.001$), consistent with greater mismatch when reported scores are high relative to the internal proxy.

Logits statistics also matter. Higher logits mean is associated with smaller distance and lower cross-channel ECE, while logits dispersion also contributes to cross-channel ECE. These effects indicate that magnitude agreement is greater when the internal proxy and reported confidence occupy more compatible ranges; they do not require treating logits-based confidence as a ground-truth target.

Model metadata are secondary after these controls. Model type, family, version, and size generally have smaller and less consistent effects across outcomes than the confidence-distribution statistics. Model size is a partial exception for correlation, where its ridge coefficient is comparable to the linguistic-distribution coefficients, but this effect does not recur consistently for distance or cross-channel ECE. Overall, the recurring predictors across outcomes are properties of the confidence distributions, whereas metadata effects are more dependent on the particular outcome and model comparison. Full coefficients and uncertainty estimates are reported in Appendix~\ref{app:regression}.

\subsection{A Lossy-Channel Interpretation}
\label{sec:lossy_channel}

These results suggest a simple interpretation. Linguistic confidence is a lossy channel from internal confidence to a user-facing score. If the output channel maps many internal states to the same verbal score, the score cannot preserve instance-level distinctions. This occurs when linguistic confidence collapses near a few high values. If the channel uses a broader range of scores, it can preserve more rank-order information, which increases association.

This interpretation also explains why dispersion is not enough. A broader score range can still be shifted upward, be nonlinear, or be poorly matched to correctness. In that case, correlation may improve but calibration can remain poor. This is why we do not claim that increasing dispersion improves alignment in general. The more precise claim is that dispersion can improve association and make confidence useful for ranking or weighting, but calibration still requires separate validation.

\section{Linguistic Confidence as a Weighting Signal}
\label{sec:aggregation}

To connect the analysis to a downstream use case, we conduct a small aggregation study with \texttt{gpt-4.1-nano} \cite{openai2025gpt41} on the eight classification tasks. For each prompt, we sample 16 outputs and aggregate predictions by voting. We compare three settings. In answer-only voting, no confidence is elicited and every prediction has weight 1. In level-confidence voting, the model reports \textit{Very confident}, \textit{Moderately confident}, or \textit{Uncertain}, which are mapped to weights 3, 2 and 1. In score-confidence voting, the model reports a numeric score from 0 to 10. For each candidate answer, we sum the weights across samples and select the answer with the highest total weight.

\begin{table}[t]
\centering
\small
\setlength{\tabcolsep}{4pt}
\begin{tabular}{lcc}
\toprule
\textbf{Method} & \textbf{Accuracy} & \textbf{Entropy} \\
\midrule
Majority voting & 0.7311 & N/A \\
Level confidence & 0.7531 & 0.4362 \\
Score confidence & 0.7830 & 1.6896 \\
\bottomrule
\end{tabular}
\caption{Confidence-weighted aggregation on eight classification tasks using \texttt{gpt-4.1-nano}. Finer-grained confidence gives a more dispersed weighting signal and improves accuracy in this setting.}
\label{tab:confidence_entropy_aggregation}
\end{table}

Table~\ref{tab:confidence_entropy_aggregation} shows that confidence-weighted aggregation improves over majority voting in this setting, and numeric confidence performs best. We interpret this as an illustrative downstream use of rank-order signal, not as evidence that linguistic confidence is calibrated. The result is consistent with our main analysis. A more diverse confidence distribution can help weight samples, but it should be validated on the target setting before deployment.

\section{Operational Guidance}
\label{sec:operational_guidance}

The empirical results imply a simple use policy as shown in Table~\ref{tab:operational_guidance}. Linguistic confidence should first be treated as a signal to be validated, not as a probability. Before using it for abstention, filtering, or sample weighting, one should check whether the reported scores have enough spread, whether they correlate with correctness or an internal proxy on held-out data, and whether high verbal confidence occurs in cases where other evidence suggests uncertainty.


\begin{table*}[t]
\centering
\small
\setlength{\tabcolsep}{5pt}
\renewcommand{\arraystretch}{1.08}
\begin{tabular}{
  @{}
  >{\raggedright\arraybackslash}p{0.24\linewidth}
  >{\raggedright\arraybackslash}p{0.34\linewidth}
  >{\raggedright\arraybackslash}p{0.34\linewidth}
  @{}
}
\toprule
\textbf{Regime} & \textbf{Diagnostic} & \textbf{Recommended use} \\
\midrule
\textbf{Collapsed verbal scores}
& Scores cluster near 9--10, or have low standard deviation
& Avoid using verbal confidence for fine-grained decisions \\
\addlinespace[0.35em]

\textbf{High verbal confidence with weak internal confidence}
& Linguistic confidence is much higher than logits-based confidence
& Flag, abstain, rerank, or sample additional answers \\
\addlinespace[0.35em]

\textbf{Dispersed verbal scores with validated association}
& Held-out data show positive association with correctness or internal confidence
& Use as a ranking or weighting signal, not as a calibrated probability \\
\addlinespace[0.35em]

\textbf{Prompt-induced confidence shifts}
& Tone or framing raises confidence without improving association
& Treat as confidence inflation, not improved reliability \\
\bottomrule
\end{tabular}
\caption{Practical interpretation of linguistic confidence under different diagnostic regimes.}
\label{tab:operational_guidance}
\end{table*}

For white-box systems, the most useful signal is often the mismatch between channels. High linguistic confidence paired with low internal confidence identifies cases where the model sounds certain despite weak probability support. For black-box systems, internal confidence is unavailable, so the safer procedure is to validate linguistic confidence on held-out examples from the same task. If reported scores are collapsed or prompt-sensitive, they should not control abstention thresholds. If they remain dispersed and validated, they can support weak ranking decisions such as confidence-weighted voting.

\section{Discussion}
\label{sec:discussion}

The main lesson is that linguistic confidence is not a transparent readout of internal uncertainty. It is a separate confidence channel with its own distributional behavior. In some regimes, it carries rank-order information. In other regimes, especially when scores collapse near high values, it becomes uninformative at the instance level. This explains why aggregate plots can look encouraging while instance-level correlations remain weak.

The distinction between association and calibration is central. A model can vary linguistic confidence in the same direction as logits-based confidence and still be systematically overconfident. This pattern appears often for instruction-tuned models. Conversely, a model can have small distance because both signals are concentrated, while still providing little useful ranking information. Reliability claims should therefore report association, magnitude agreement and calibration together.

The results also clarify what prompt design can and cannot do. Prompting can alter the distribution of reported confidence, and careful exemplar choices can avoid complete score collapse. This may help when confidence is used as a weak ranking signal. But prompt tone can inflate confidence without improving alignment. A prompt that asks the model to reflect its internal probabilities does not by itself make verbal confidence track those probabilities.

Our analysis is diagnostic rather than causal. We do not claim to identify the training changes that create a given confidence distribution, and we do not claim that manipulating variance alone will fix confidence. The evidence shows that many alignment patterns attributed to family, size, or instruction tuning are better understood through the mean, variance and saturation of the two confidence channels. This is the main reason to evaluate linguistic confidence as a channel with measurable failure modes, rather than as a direct probability estimate.

\section{Conclusion}

We study whether linguistic confidence aligns with internal confidence signals in LLMs. Across 10 tasks and 30 models, instance-level alignment is weak. Association tends to be higher on easier tasks and for stronger base models, but instruction-tuned models often remain overconfident. Prompting can change reported confidence distributions, but attitude cues inflate confidence without improving alignment. Regression analyses show that distributional properties explain much of the observed pattern. Linguistic dispersion can preserve rank-order signal, but it does not guarantee magnitude agreement or calibration.

Linguistic confidence is best treated as a lossy confidence channel. It can be useful for downstream weighting when validated, but it should not be treated as a calibrated probability. We recommend reporting multi-axis confidence diagnostics, checking for collapsed or inflated confidence distributions, and using linguistic confidence only with distribution-aware validation or held-out calibration.

\section*{Limitations}

Our study focuses on a limited set of open-source models and two internal confidence proxies. Logits-based confidence is not ground-truth uncertainty, and semantic entropy is also an estimate with modeling choices such as sampling and NLI-based clustering. Our goal is to compare available confidence channels, not to identify true uncertainty.

Most experiments are correlational. The regression analysis controls for observable factors such as confidence mean, variance, task structure and performance, but it does not establish a causal mechanism. We also do not have access to the training details needed to explain why a particular model version changes its confidence distribution.

The repeated-subsampling analysis reduces concern that the main trends are artifacts of a particular 200-example subset, but individual estimates still have sampling variability. The black-box aggregation study is small, uses one model outside the main suite, and cannot evaluate the paper's cross-channel diagnostics; it should be read only as an illustration.

Future work should compare additional internal uncertainty proxies, especially for generation, and test whether conclusions are stable under alternative proxy choices. Another important direction is post-hoc confidence elicitation: first obtain the model's answer, then ask it to assess confidence in that completed answer. This protocol may reveal whether reflection changes expressed confidence, but it requires a separate generation setting and is outside the scope of the present experiments.

\section*{Ethical Considerations}

This work studies when user-facing confidence statements from LLMs can be misleading. The main ethical risk is that confidence scores may create unwarranted trust, especially in settings where users cannot inspect internal uncertainty signals. Our results suggest that high linguistic confidence should not be presented as a calibrated probability unless it has been validated for the specific task and deployment context.

The study uses existing NLP benchmarks and publicly available models, with no human-subject data collection. The practical recommendations are intended to reduce overreliance on unsupported confidence claims by encouraging multi-axis reporting, mismatch detection and held-out calibration. However, these diagnostics can also be misused if presented as guarantees of reliability. We therefore caution against using linguistic confidence as the sole basis for decisions in high-stakes applications.

\section*{Use of AI Tools}
AI tools were used solely for final-stage proofreading and formatting checks, including grammar correction and verification of LaTeX compilation and document formatting. They were not used to generate research ideas, conduct analyses, or produce substantive content for the manuscript.

\section*{Acknowledgement}
This research was supported in part by the National Science Foundation under Grant No. 2452367.

\bibliography{custom}
\clearpage
\appendix

\section{Prompt Design and Robustness Analyses}
\label{appendix:prompt}
This appendix provides detailed prompting setups and additional robustness analyses omitted from the main paper. We describe the confidence elicitation prompts used for both classification and generation tasks, followed by analyses of prompt sensitivity, including few-shot confidence exemplar search, confidence-elicitation ablation, and attitude-based prompt perturbations.
\subsection{Classification Prompt Design}

For classification tasks, we design prompts to elicit linguistic confidence while maintaining consistent output formats across model types. We employ two prompt templates for base models. Prompt 1 is adapted from \citet{xiong2023can} and directly elicits linguistic confidence alongside the answer. Prompt 2 additionally explains the meaning of confidence scores before elicitation. For instruction-tuned models, confidence reporting and format constraints are specified through the system prompt.

We explore multiple confidence exemplar settings and select $(5,6)$ for Prompt 1 and $(4,5)$ for Prompt 2 on base models, while score 2 is used as the format example for instruction-tuned models. The rationale and quantitative comparison of different exemplar scores are reported in Appendix~\ref{appendix:hypersearch}.

\begin{figure}[h]
  \includegraphics[width=\linewidth]{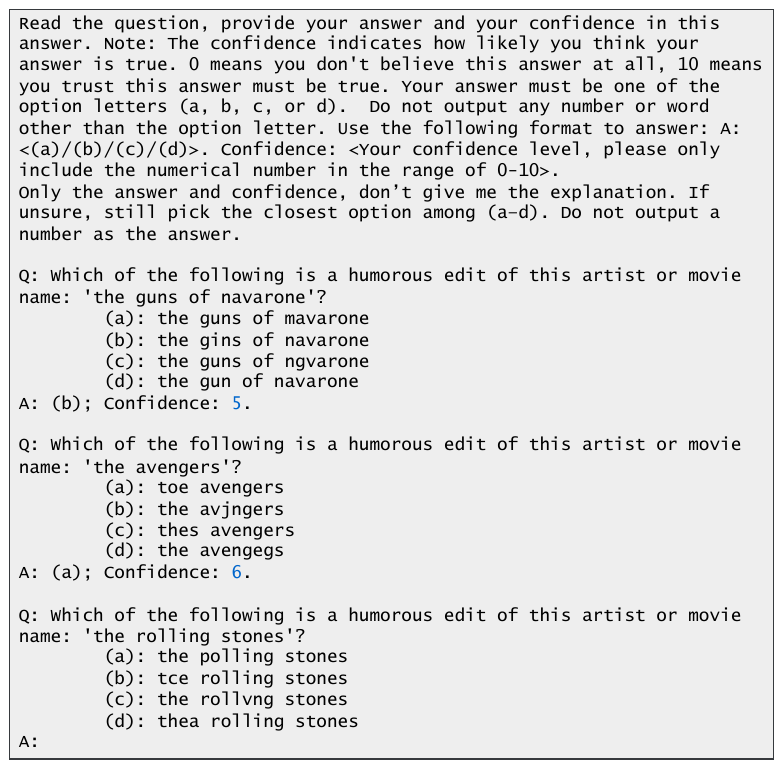}
  \caption {Base Prompt 1 example.}
    \label{fig:prompt1_base}
\end{figure}

\begin{figure}[h]
  \includegraphics[width=\linewidth]{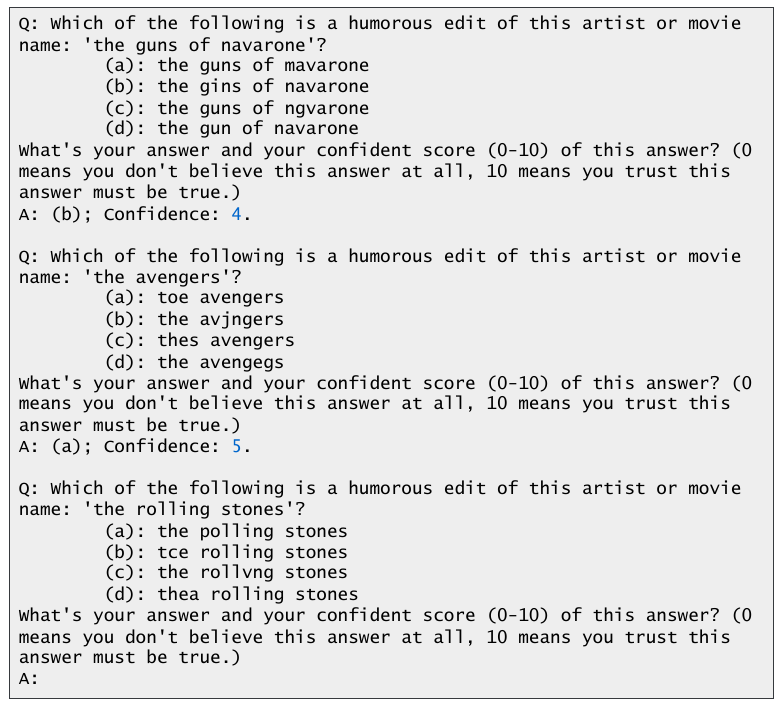}
  \caption {Base Prompt 2 example.}
    \label{fig:prompt2_base}
\end{figure}

\begin{figure}[h]
  \includegraphics[width=\linewidth]{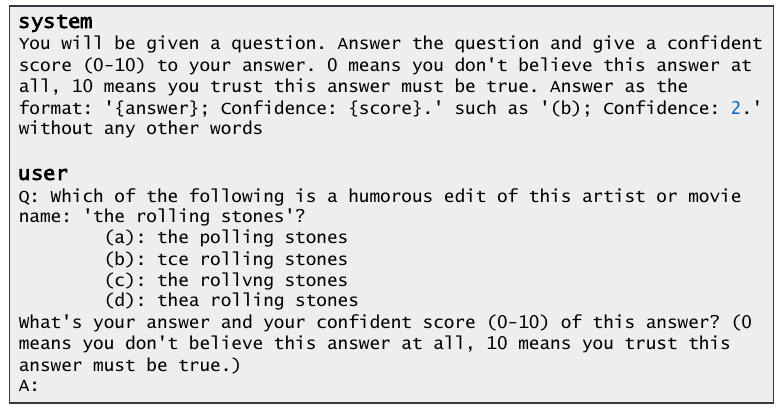}
  \caption {Instruct prompt with example confidence score.}
    \label{fig:prompt_instruct}
\end{figure}

\begin{figure}[h]
  \includegraphics[width=\linewidth]{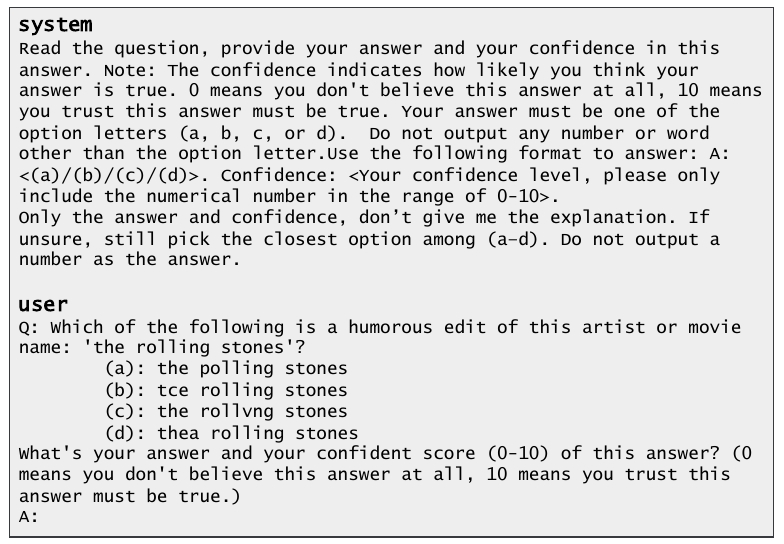}
  \caption {Instruct prompt without example confidence score.}
    \label{fig:prompt_instruct_noscore}
\end{figure}

Prompt examples on the \textit{Ruin Names} task are shown in Figures~\ref{fig:prompt1_base}, \ref{fig:prompt2_base}, \ref{fig:prompt_instruct}, and \ref{fig:prompt_instruct_noscore}.

\subsection{Generation Prompt Design}

For generation tasks, we follow the prompting setup of \citet{kuhn2023semantic} and append linguistic confidence elicitation instructions following \citet{xiong2023can}. We evaluate two representative generation benchmarks, \textit{CoQA} and \textit{TriviaQA}. 

For CoQA, each instance consists of a context paragraph and several question-answer pairs, where we adopt a zero-shot prompting setup. For TriviaQA, we employ a ten-shot prompting strategy to improve response accuracy. In both settings, linguistic confidence is elicited alongside generated responses and averaged across the five sampled generations used for semantic-entropy estimation.

\begin{figure}[t]
  \includegraphics[width=\linewidth]{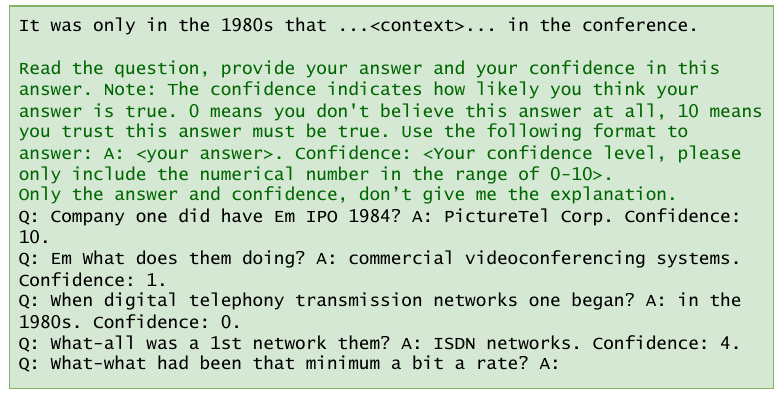}
  \caption {Example prompt for CoQA.}
    \label{fig:coqa}
\end{figure}

\begin{figure}[t]
  \includegraphics[width=\linewidth]{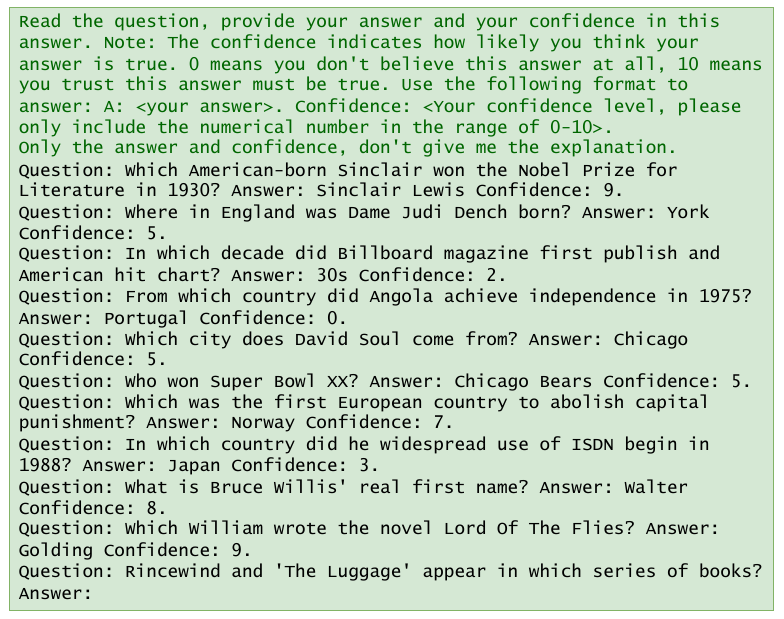}
  \caption {Example prompt for TriviaQA.}
    \label{fig:triviaqa}
\end{figure}

Representative prompts are shown in Figures~\ref{fig:coqa} and \ref{fig:triviaqa}.

\subsection{Confidence-Elicitation Ablation}
\label{appendix:ablation_prompt}

To examine whether confidence elicitation itself perturbs internal probability distributions, we conduct ablation experiments removing linguistic confidence instructions while keeping all other prompt settings unchanged. For base models, we retain the original two-shot examples without confidence elicitation, while instruction-tuned models preserve the same formatting constraints.

\begin{figure}[t]
  \includegraphics[width=\linewidth]{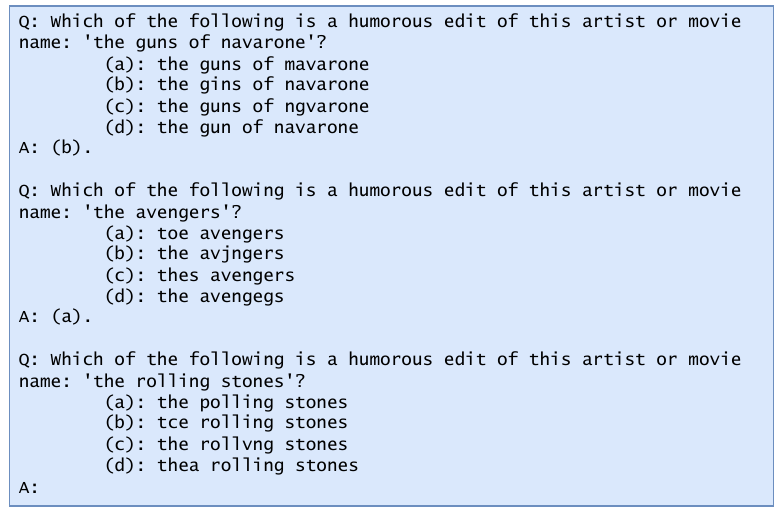}
  \caption {Base Prompt without asking for linguistic confidence.}
    \label{fig:base_withoutconf}
\end{figure}

\begin{figure}[t]
  \includegraphics[width=\linewidth]{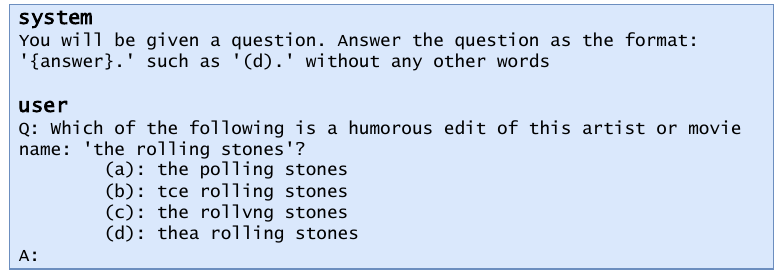}
  \caption {Instruct prompt without asking for linguistic confidence.}
    \label{fig:instruct_withoutconf}
\end{figure}

Representative prompts without confidence elicitation are shown in Figures~\ref{fig:base_withoutconf} and \ref{fig:instruct_withoutconf}. Quantitative comparisons of accuracy, logits statistics, and calibration are reported in Appendix~\ref{appendix:ablation}.

\subsection{Attitude-Based Prompt Perturbation}

To investigate whether user attitude affects reported confidence, we prepend criticism, approval, and irrelevant cues before the original task prompt. Each attitude category contains three cue variants. These perturbations are designed to test whether linguistic confidence reflects underlying model uncertainty or is sensitive to conversational framing.

The attitude cues used in our experiments are shown in Figure~\ref{fig:attitude}. Full quantitative results are reported in Appendix~\ref{appendix:prompt_perturbation}.
\begin{figure}[t]
  \includegraphics[width=\linewidth]{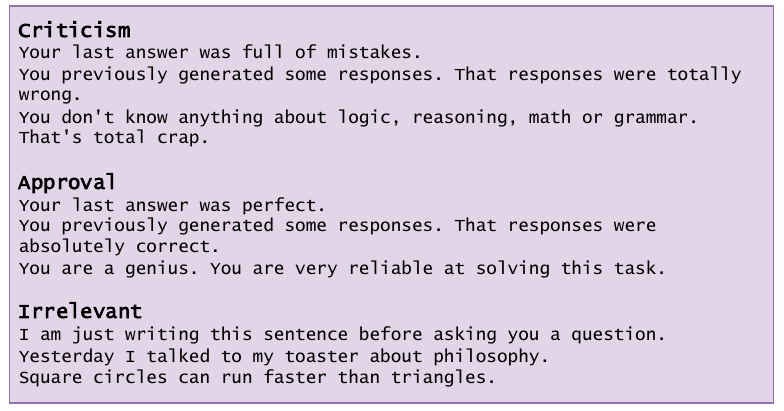}
  \caption {Attitude cues for prompt perturbation.}
    \label{fig:attitude}
\end{figure}

\section{Dataset Stability Analysis}
\label{app:dataset_stability}

Downsampling is used for computational tractability only. To test whether the 200-example limit systematically changes the results, we construct larger evaluation pools for the five affected tasks and evaluate nine model--prompt settings completed for every task. For CoLA and Temporal Sequences, we use the full evaluation sets; for QNLI and MMLU, we use substantially enlarged evaluation pools. Because the QQP evaluation set is considerably larger and expensive to evaluate in full, we randomly sample 5,000 examples as a proxy full set for the stability analysis. For each task and setting, we draw 500 random subsamples of 200 examples from the corresponding larger pool and recompute accuracy, normalized accuracy, Pearson, Spearman, and Kendall correlations, ECE, and cross-channel distance. The exact main-paper subsets, the QQP proxy set, and the resampling code are included in the released repository.

Across tasks, the median absolute difference between a subsample estimate and its corresponding larger-pool estimate is 0.024--0.027, and 93.9\%--95.3\% of estimates fall within $\pm 0.10$. The larger-pool estimate falls within the 95\% interval of the subsampling distribution for 581 of 585 metric rows overall and for all 225 rows involving the five metrics most directly tied to our conclusions. Thus, the 200-example evaluations recover the trends observed in the larger evaluation pools without detectable systematic distortion, although individual estimates retain sampling variability.
\begin{table*}[t]
\centering
\small
\resizebox{\textwidth}{!}{
\begin{tabular}{lrrrrr}
\toprule
Task / pool & Pool $n$ & Median $|\Delta|$ & Within $\pm0.10$ &
Full in 95\% interval & Conclusion-level metrics \\
\midrule
CoLA            & 1,043 & 0.024 & 95.3\% & 117/117 & 45/45 \\
QNLI            & 5,463 & 0.027 & 93.9\% & 116/117 & 45/45 \\
QQP             & 5,000 & 0.027 & 94.1\% & 115/117 & 45/45 \\
MMLU            & 1,000 & 0.026 & 94.4\% & 116/117 & 45/45 \\
Temporal Sequences        &   800 & 0.024 & 95.3\% & 117/117 & 45/45 \\
\bottomrule
\end{tabular}
}
\caption{Stability of estimates from 500 random 200-example subsamples.}
\label{tab:subsampling_stability}
\end{table*}

\section{Model Selection}
\label{appendix:model_selection}
We select 30 models from the LLaMA, Mistral and Qwen families. The full list is shown in Table~\ref{tab:model_seletion}. For the small-scale aggregation study, we use \texttt{gpt-4.1-nano} through the OpenAI API, accessed May 4, 2026, with temperature 0.7.

\begin{table}[t]
\centering
\small
\begin{tabular}{ll}
\toprule
\textbf{Base Model} & \textbf{Instruction-tuned / Chat} \\
\midrule
Llama-2-7b-hf & Llama-2-7b-chat-hf \\
Llama-2-13b-hf & Llama-2-13b-chat-hf \\
 Llama-3.1-8B & Meta-Llama-3.1-8B-Instruct \\
Llama-3.2-3B & Llama-3.2-3B-Instruct \\
Meta-Llama-3-8B & Meta-Llama-3-8B-Instruct \\
\midrule
Mistral-7B-v0.1 & Mistral-7B-Instruct-v0.1 \\
Mistral-7B-v0.2 & Mistral-7B-Instruct-v0.2 \\
Mistral-7B-v0.3 & Mistral-7B-Instruct-v0.3 \\
Mistral-Nemo-Base-2407 & Mistral-Nemo-Instruct-2407 \\
Mixtral-8x7B-v0.1 & Mixtral-8x7B-Instruct-v0.1 \\
\midrule
Qwen2.5-3B & Qwen2.5-3B-Instruct \\
Qwen2.5-7B & Qwen2.5-7B-Instruct \\
Qwen2.5-14B & Qwen2.5-14B-Instruct \\
Qwen3-4B-Base & Qwen3-4B-Instruct-2507 \\
Qwen3-8B-Base & -- \\
Qwen3-14B-Base & -- \\
\bottomrule
\end{tabular}
\caption{List of models used in our experiments, categorized by version and type.}
\label{tab:model_seletion}
\end{table}

\section{Distance and ECE trends by Model Size}
\label{appendix:trends_by_size}
\begin{figure}[htbp]
    \centering
    \begin{subfigure}[t]{0.235\textwidth}
        \centering
        \includegraphics[width=\linewidth]{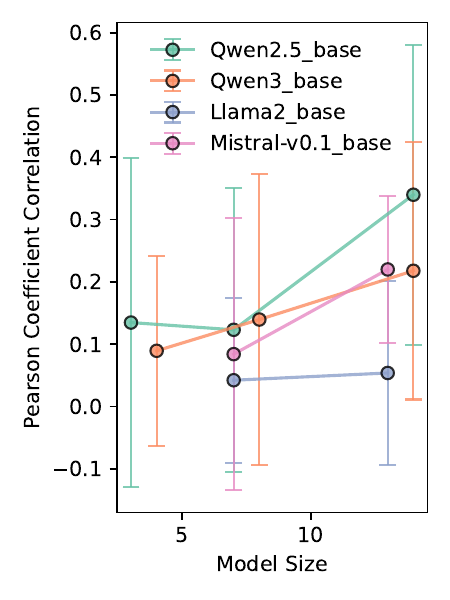}
        \caption{Base Model.}
        \label{fig:left}
    \end{subfigure}
    \begin{subfigure}[t]{0.235\textwidth}
        \centering
        \includegraphics[width=\linewidth]{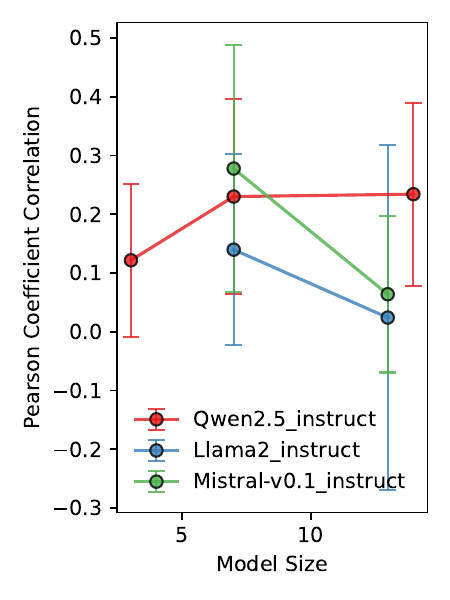}
        \caption{Instruction-tuned Model.}
        \label{fig:right}
    \end{subfigure}

    \caption{Effect of model size on the correlation between linguistic and logits-based confidence, with error bars indicating the standard deviation of correlation values within each model group. Correlation tends to rise with model size for base models but plateaus for instruction-tuned ones.}
    \label{fig:corrbymodel_withbar}
\end{figure}
\begin{figure*}[t]
  \centering
  \begin{subfigure}[t]{0.33\textwidth}
    \centering
    \includegraphics[width=\linewidth]{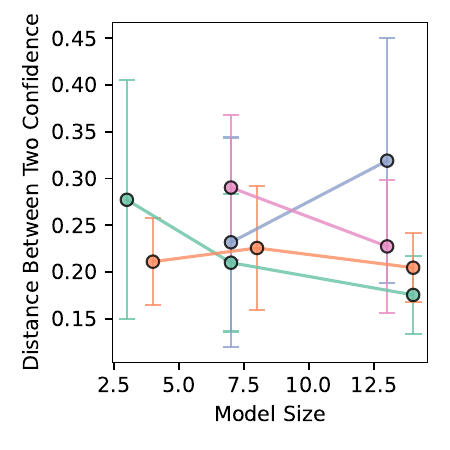}
    \caption{Distance}
    \label{fig:base-dist}
  \end{subfigure}\hfill
  \begin{subfigure}[t]{0.33\textwidth}
    \centering
    \includegraphics[width=\linewidth]{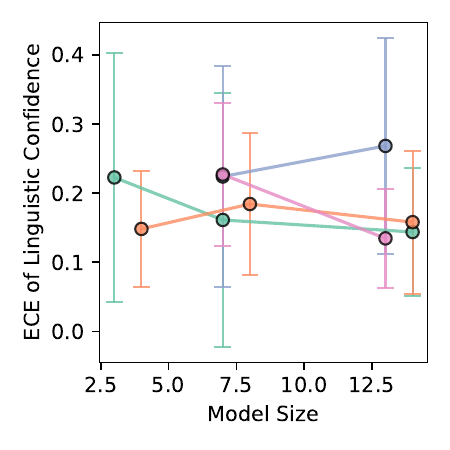}
    \caption{ECE(linguistic)}
    \label{fig:base-ece-li}
  \end{subfigure}\hfill
  \begin{subfigure}[t]{0.27\textwidth}
    \centering
    \raisebox{1cm}{\includegraphics[width=.9\linewidth]{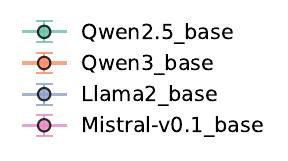}}
    \label{fig:legend-base}
  \end{subfigure}

  \caption{Base models: model size (x-axis, in B) vs. (a) Distance between two confidence measures and (b) ECE of linguistic confidence.}
  \label{fig:base-all}
\end{figure*}
\begin{figure*}[t]
  \centering
  \begin{subfigure}[t]{0.33\textwidth}
    \centering
    \includegraphics[width=\linewidth]{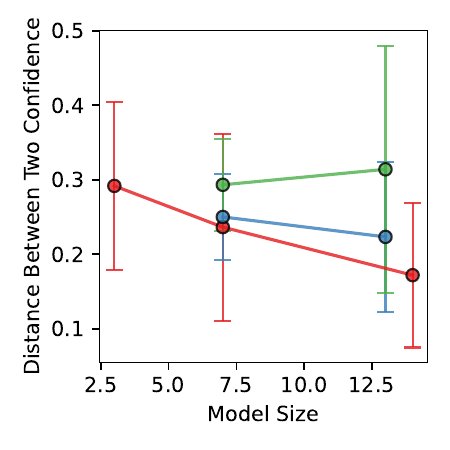}
    \caption{Distance}
    \label{fig:inst-dist}
  \end{subfigure}\hfill
  \begin{subfigure}[t]{0.33\textwidth}
    \centering
    \includegraphics[width=\linewidth]{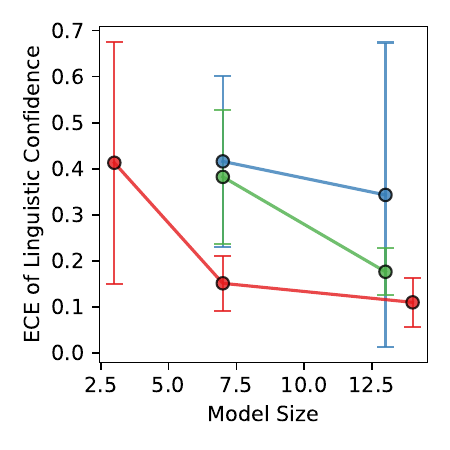}
    \caption{ECE(linguistic)}
    \label{fig:inst-ece-li}
  \end{subfigure}\hfill
  \begin{subfigure}[t]{0.27\textwidth}
    \centering
    \raisebox{1.2cm}{\includegraphics[width=.9\linewidth]{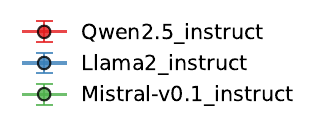}}
    \label{fig:legend-instruct}
  \end{subfigure}

  \caption{Instruction-tuned models: model size (x-axis, in B) vs. (a) Distance between two confidence measures and (b) ECE of linguistic confidence.}
  \label{fig:inst-all}
\end{figure*}

For completeness, we present the error-bar version of the model-size correlation plot in Figure~\ref{fig:corrbymodel_withbar}. The error bars indicate the standard deviation of correlation values within each model group. Figures~\ref{fig:base-all} and~\ref{fig:inst-all} show how model size affects distance and ECE for base and instruction-tuned models. These plots illustrate why one metric cannot replace the others. Base models often show higher correlation with scale, while distance and ECE may move differently. Instruction-tuned models can also show lower distance or ECE as accuracy improves, even when correlation does not improve. This supports our main claim that association, magnitude agreement and calibration measure different aspects of confidence reliability.

\section{Distribution of Logits-based and Linguistic Confidence}
\label{appendix:logits_conf_distribution}
\begin{figure}[t]
  \centering
  \begin{subfigure}[t]{0.48\textwidth}
    \centering
    \includegraphics[width=\linewidth]{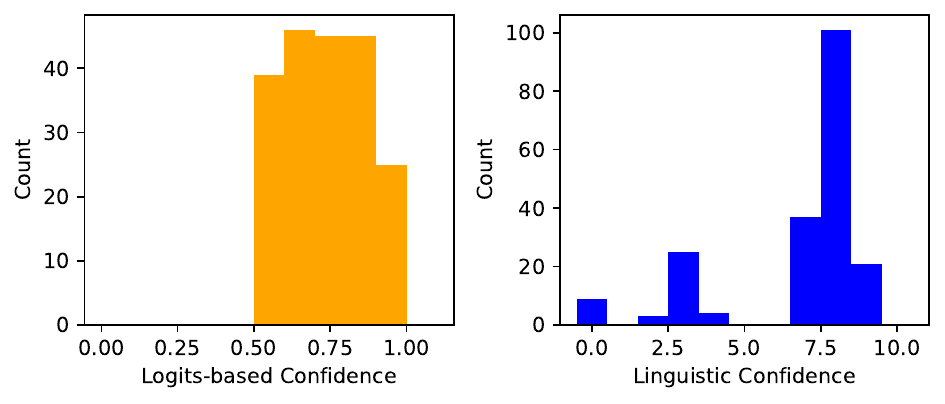}
    \caption{Qwen2.5-3B}
  \end{subfigure}\hfill
  \begin{subfigure}[t]{0.48\textwidth}
    \centering
    \includegraphics[width=\linewidth]{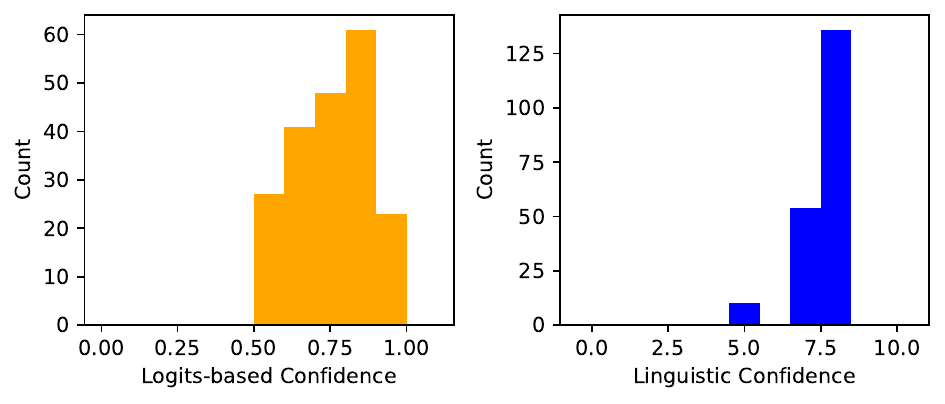}
    \caption{Qwen2.5-7B}
  \end{subfigure}\hfill
  \begin{subfigure}[t]{0.48\textwidth}
    \centering
    \includegraphics[width=\linewidth]{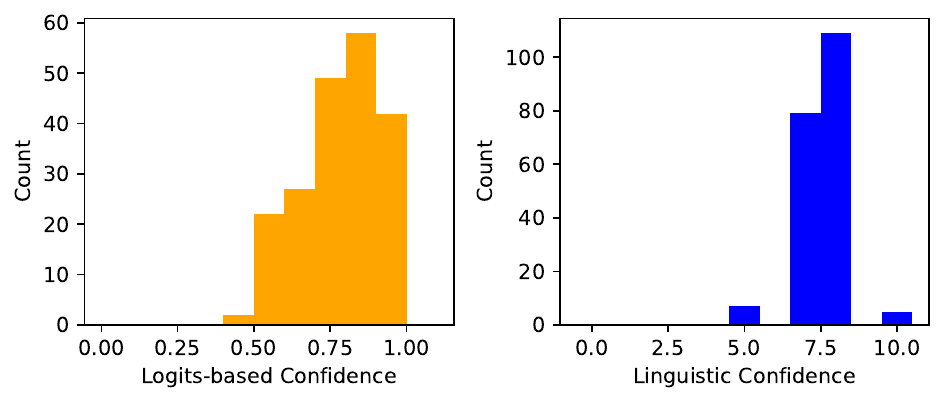}
    \caption{Qwen2.5-14B}
  \end{subfigure}\hfill
  \caption{Distribution of Logits-based and Linguistic Confidence on CoLA (Base Prompt 1) for Qwen2.5 Base Models.}
  \label{fig:qwen2.5basecola}
\end{figure}

\begin{figure}[t]
  \centering
  \begin{subfigure}[t]{0.48\textwidth}
    \centering
    \includegraphics[width=\linewidth]{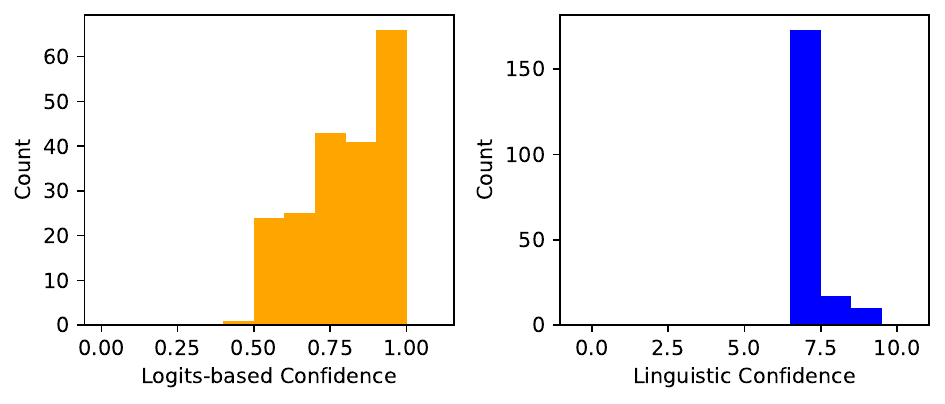}
    \caption{Qwen3-4B}
  \end{subfigure}\hfill
  \begin{subfigure}[t]{0.48\textwidth}
    \centering
    \includegraphics[width=\linewidth]{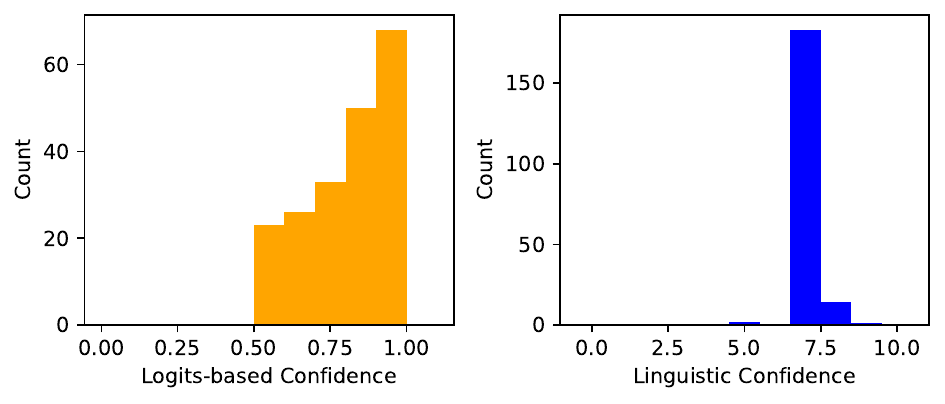}
    \caption{Qwen3-8B}
  \end{subfigure}\hfill
  \begin{subfigure}[t]{0.48\textwidth}
    \centering
    \includegraphics[width=\linewidth]{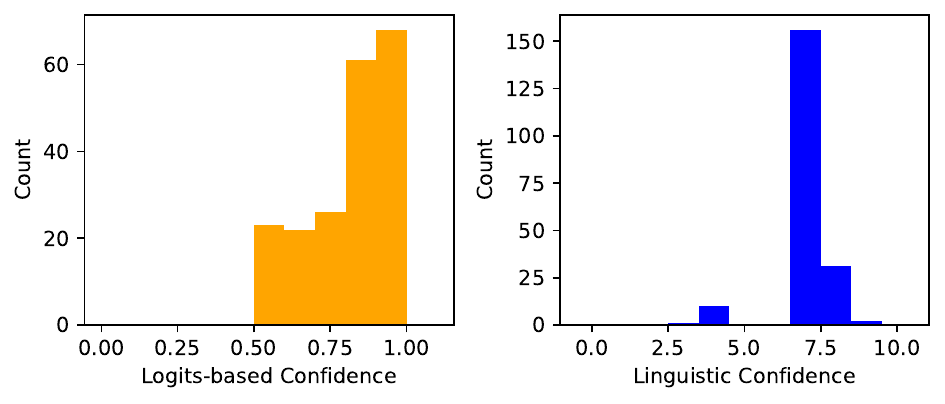}
    \caption{Qwen3-14B}
  \end{subfigure}\hfill
  \caption{Distribution of Logits-based and Linguistic Confidence on CoLA (Base Prompt 1) for Qwen3 Base Models.}
  \label{fig:qwen3basecola}
\end{figure}

\begin{figure}[t]
  \centering
  \begin{subfigure}[t]{0.48\textwidth}
    \centering
    \includegraphics[width=\linewidth]{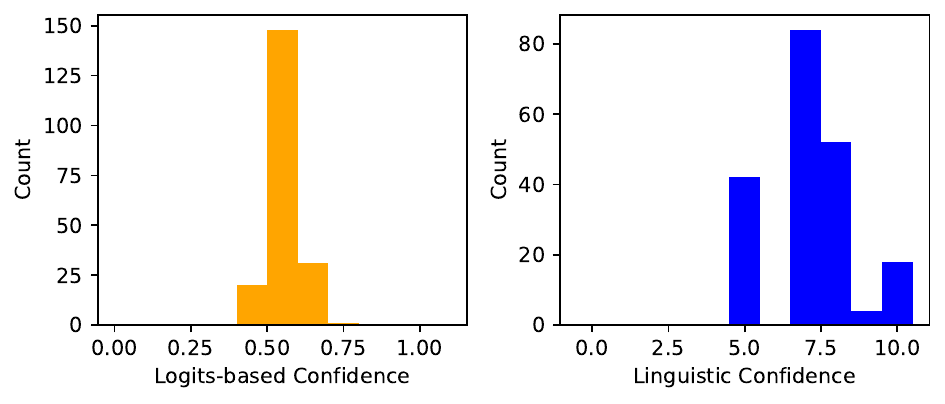}
    \caption{Llama-2-7b-hf}
  \end{subfigure}\hfill
  \begin{subfigure}[t]{0.48\textwidth}
    \centering
    \includegraphics[width=\linewidth]{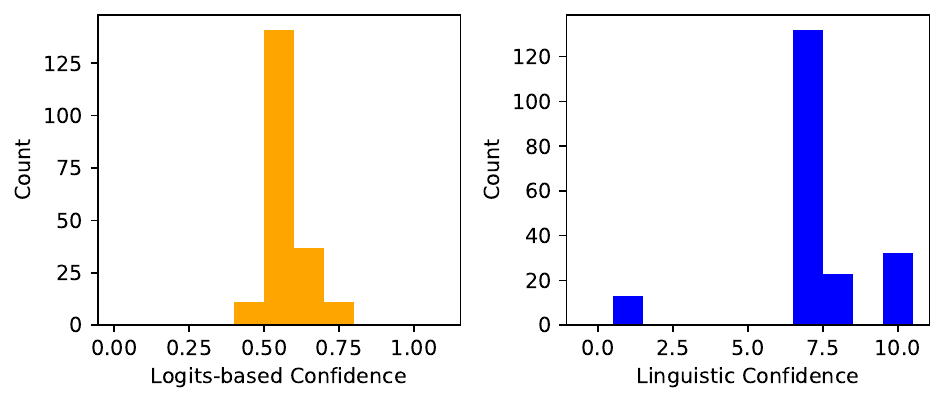}
    \caption{Llama-2-13b-hf}
  \end{subfigure}\hfill
  \caption{Distribution of Logits-based and Linguistic Confidence on CoLA (Base Prompt 1) for Llama2 Base Models.}
  \label{fig:llama2basecola}
\end{figure}

As discussed in Section~\ref{Chapter:class_corr_model}, the correlation generally increases with model size for base models, but decreases for instruction-tuned models.
To illustrate this pattern in detail, we examine the \textit{CoLA} task. Besides Figure~\ref{fig:mistral_cola_distributions}, Qwen and Llama show the similar pattern. As shown in Figures~\ref{fig:qwen3basecola} and \ref{fig:llama2basecola}, linguistic confidence in base models exhibits greater diversity and a less concentrated distribution.
An exception occurs with \texttt{Qwen2.5} (Figure~\ref{fig:qwen2.5basecola}), where the 7B model produces more monotonous linguistic confidence compared to the 3B model, consistent with the overall trend observed in Figure~\ref{fig:corrbymodel}.

\begin{figure}[t]
  \centering
  \begin{subfigure}[t]{0.48\textwidth}
    \centering
    \includegraphics[width=\linewidth]{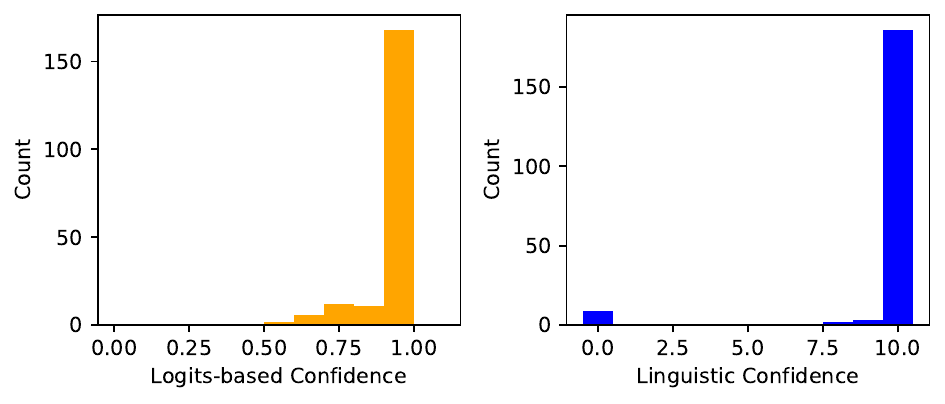}
    \caption{Qwen2.5-3B-Instruct}
  \end{subfigure}\hfill
  \begin{subfigure}[t]{0.48\textwidth}
    \centering
    \includegraphics[width=\linewidth]{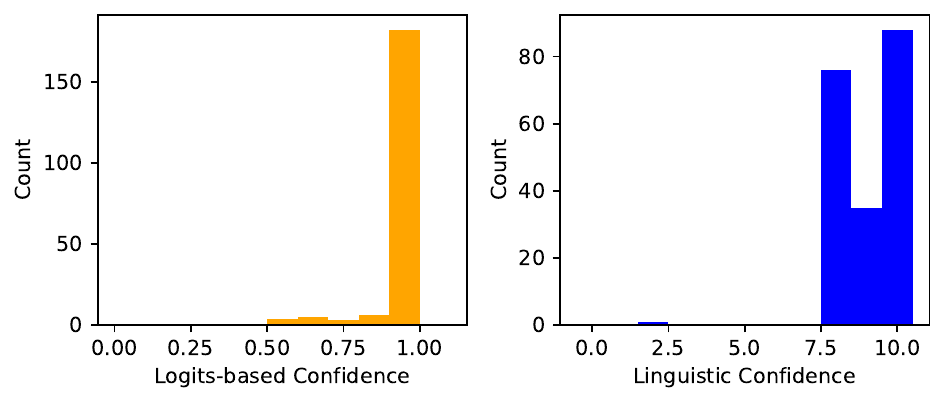}
    \caption{Qwen2.5-7B-Instruct}
  \end{subfigure}\hfill
  \begin{subfigure}[t]{0.48\textwidth}
    \centering
    \includegraphics[width=\linewidth]{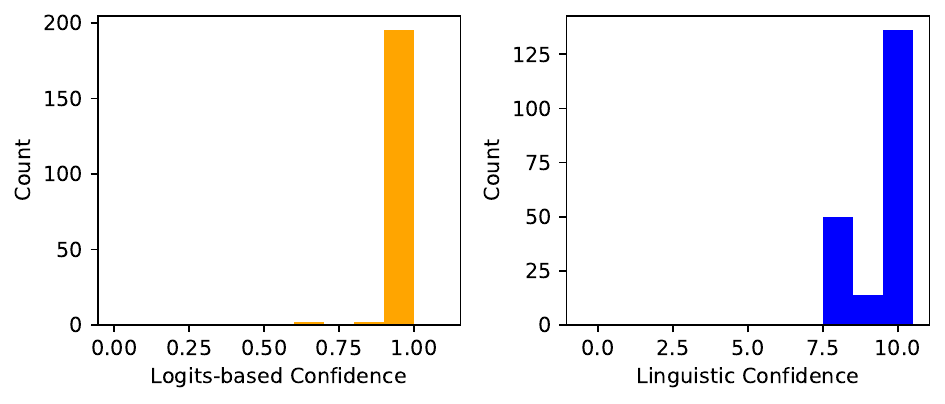}
    \caption{Qwen2.5-14B-Instruct}
  \end{subfigure}\hfill
  \caption{Distribution of Logits-based and Linguistic Confidence on CoLA for Qwen2.5 Instruct-tuned Models.}
  \label{fig:qwen2.5inscola}
\end{figure}

\begin{figure}[t]
  \centering
  \begin{subfigure}[t]{0.48\textwidth}
    \centering
    \includegraphics[width=\linewidth]{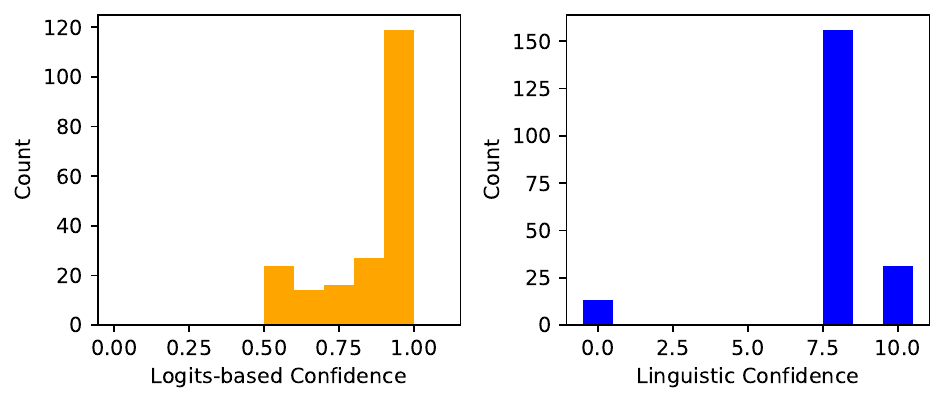}
    \caption{Llama-2-7b-chat-hf}
  \end{subfigure}\hfill
  \begin{subfigure}[t]{0.48\textwidth}
    \centering
    \includegraphics[width=\linewidth]{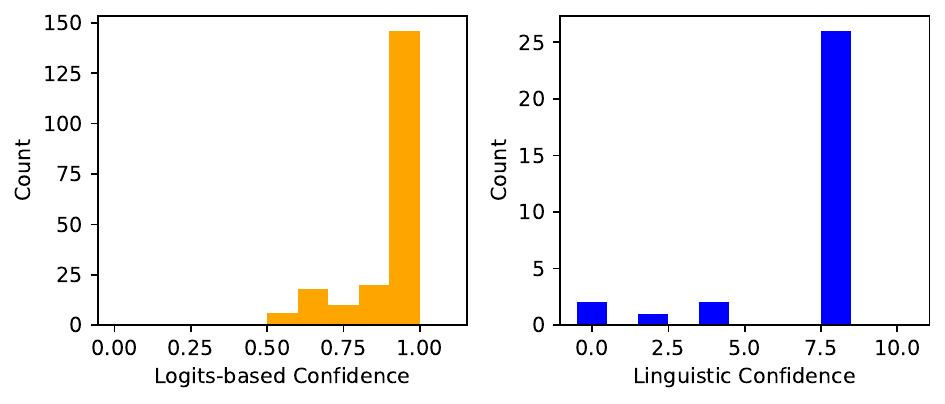}
    \caption{Llama-2-13b-chat-hf}
  \end{subfigure}\hfill
  \caption{Distribution of Logits-based and Linguistic Confidence on CoLA for Llama2 Instruction-tuned Models.}
  \label{fig:llama2inscola}
\end{figure}

For instruction-tuned models (Figures~\ref{fig:qwen2.5inscola} and \ref{fig:llama2inscola}), the logits-based confidence values become larger and often approach 1 as model size increases.
This saturation of logits leads to a reduction in the correlation between the two confidence measures.

\section{Distribution of Correlation Scores and Paired t-Test Analyses}
\label{appendix:ttest_family_version}
To further examine how model confidence alignment evolves across different architectures and training versions, 
we conducted paired $t$-tests on both \textbf{model families} and \textbf{model versions}. 
For the family-level comparison, we selected representative base models from the LLaMA, Mistral, and Qwen series, 
and evaluated whether their overall correlation, confidence-logit distance, and ECE differ significantly 
(\autoref{tab:family_ttest}). 
For the version-level comparison, we tested multiple generations within each family 
(e.g., LLaMA3$\rightarrow$LLaMA3.1, Mistral-v0.1$\rightarrow$v0.2$\rightarrow$v0.3, and Qwen2.5$\rightarrow$Qwen3), 
to investigate whether improvements in alignment and calibration persist as the models evolve 
(\autoref{tab:version_ttest}).
\begin{table*}[!ht]
\centering
\small
\begin{tabular}{lcccccc}
\toprule
\textbf{Metric} & \textbf{LLaMA} & \textbf{Mistral} & \textbf{Qwen} &
$p_{L\rightarrow M}$ & $p_{M\rightarrow Q}$ & $p_{L\rightarrow Q}$ \\
\midrule
$r$     & 0.081 & 0.137 & 0.180 & 0.033 & 0.102 & \underline{0.0004} \\
Distance & 0.274 & 0.259 & 0.218 & 0.287 & \underline{0.0009} & \underline{0.00002} \\
ECE(linguistic)         & 0.261 & 0.219 & 0.186 & \underline{0.024} & 0.070 & \underline{0.0003} \\
\bottomrule
\end{tabular}
\caption{Paired $t$-tests across model families. Mean value of the three families and $p$ between them are reported. Underlined values indicate significant differences ($p<0.05$).}
\label{tab:family_ttest}
\end{table*}

\begin{table}[!ht]
\centering
\small
\begin{tabular}{lccc}
\toprule
\textbf{Metric} & \textbf{mean\_a} & \textbf{mean\_b} & $p$ \\
\midrule
\multicolumn{4}{l}{\textit{LLaMA3 $\rightarrow$ LLaMA3.1}} \\
$r$ & 0.057 & 0.126 & \underline{0.028} \\
Distance & 0.272 & 0.281 & 0.527 \\
ECE (linguistic) & 0.197 & 0.234 & 0.058 \\
\midrule
\multicolumn{4}{l}{\textit{Mistral-v0.1 $\rightarrow$ v0.2}} \\
$r$ & 0.125 & 0.116 & 0.741 \\
Distance & 0.291 & 0.261 & 0.088 \\
ECE (linguistic) & 0.258 & 0.204 & \underline{0.011} \\
\midrule
\multicolumn{4}{l}{\textit{Mistral-v0.2 $\rightarrow$ v0.3}} \\
$r$ & 0.114 & 0.129 & 0.570 \\
Distance & 0.261 & 0.271 & 0.495 \\
ECE (linguistic) & 0.204 & 0.209 & 0.537 \\
\midrule
\multicolumn{4}{l}{\textit{Qwen2.5 $\rightarrow$ Qwen3}} \\
$r$ & 0.340 & 0.218 & \underline{0.00064} \\
Distance & 0.175 & 0.205 & \underline{0.00038} \\
ECE (linguistic) & 0.144 & 0.151 & 0.261 \\
\bottomrule
\end{tabular}
\caption{Paired $t$-tests across model versions within each family. 
Underlined values indicate significant differences ($p<0.05$). Metrics include overall correlation ($r$), distance between linguistic and logits-based confidence, and ECE based on linguistic confidence.}
\label{tab:version_ttest}
\end{table}

Across model families (Table~\ref{tab:family_ttest}), we observe a general increase in correlation from LLaMA to Mistral and Qwen, together with lower distance and ECE. This suggests stronger aggregate alignment for Qwen in our setting. Within-family comparisons in Table~\ref{tab:version_ttest} are less uniform. LLaMA3.1 improves correlation over LLaMA3, while Qwen3 shows lower correlation and larger distance than Qwen2.5. The Mistral versions show only small changes. 

We analyze the effect of model updates on the distributional properties of linguistic confidence and find that dispersion plays a central role in determining alignment. Specifically, linguistic-confidence standard deviation increases from $1.16$ to $1.32$ for the 8B Llama base models. In contrast, Qwen3 exhibits a substantial collapse in dispersion relative to Qwen2.5, with standard deviation decreasing from $1.23$ to $0.71$ for the 14B base models. This contrast aligns with their divergent correlation trends, where LLaMA models show improved alignment and Qwen models degrade. Higher dispersion can preserve instance-level rank information, whereas compressed distributions reduce expressivity and weaken association. While we do not make causal claims, these differences may be related to broader shifts in training and post-training objectives across model updates. For example, LLaMA3.1 introduces changes in data scale and alignment procedures \cite{grattafiori2024llama}, while newer models such as Qwen3 place increasing emphasis on reasoning and structured generation \cite{yang2025qwen3technicalreport}. Such changes may influence how confidence is expressed in language, potentially affecting its dispersion. However, establishing a direct causal link between training strategies and confidence distribution remains an important direction for future work. These results suggest that model identity alone does not explain alignment. Distributional changes in the confidence scores account for much of the observed pattern.

\section{Effects of Prompt Design}
\label{appendix:effect_of_prompt}
\subsection{Hyperparameter Search}
\label{appendix:hypersearch}
\begin{figure}[htbp] 
    \centering
    \begin{subfigure}{0.9\linewidth}
        \centering
        \includegraphics[width=\linewidth]{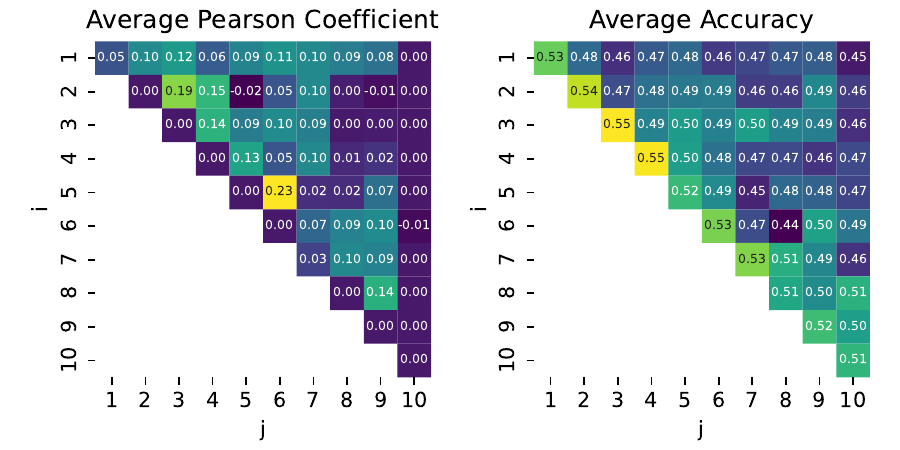}
        \caption{Prompt 1}
        \label{fig:top}
    \end{subfigure}
    \vspace{0.5em} 

    \begin{subfigure}{0.9\linewidth}
        \centering
        \includegraphics[width=\linewidth]{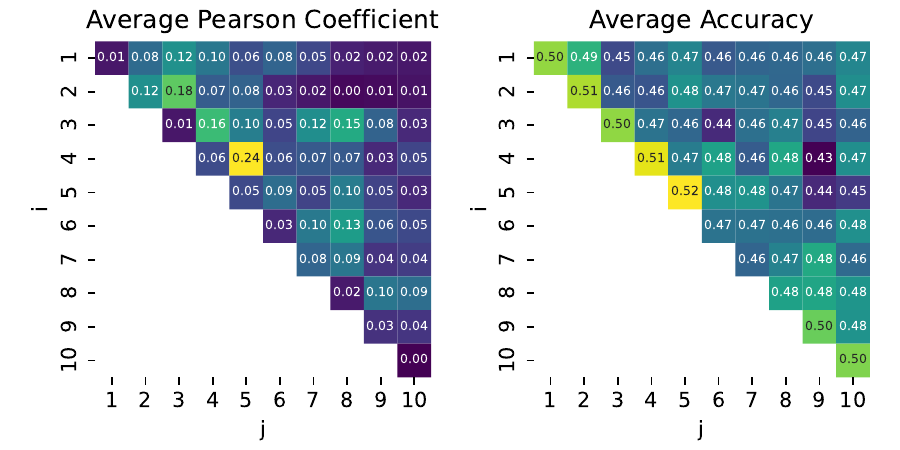}
        \caption{Prompt 2}
        \label{fig:bottom}
    \end{subfigure}

    \caption{Comparison of average Pearson correlation and accuracy across different confidence score pairs $(i, j)$ on two base prompt settings. Each heatmap shows how the model's linguistic confidence and task performance vary with score combinations.}

    \label{fig:heatmap_base}
\end{figure}

\begin{figure} [htbp]
       \includegraphics[width=\linewidth]{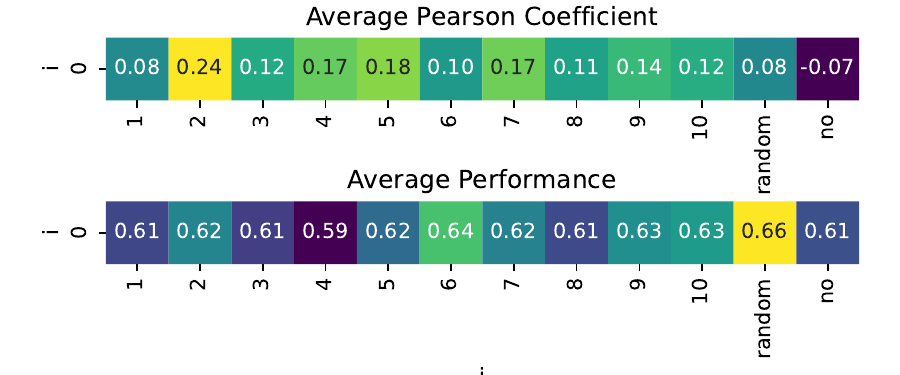}
    \caption{Average correlation (top) and task performance (bottom) across different confidence scoring schemes. 
The $x$-axis indicates the index of linguistic confidence scores (1--10), a random choice, or no confidence used. }

\label{fig:heatmap_instruct}

\end{figure}

We exhaustively explore few-shot example confidence scores to quantify how exemplar settings affect both the association between linguistic and internal confidence and task performance. Experiments cover three tasks (Conceptual Combinations, Ruin Names, Temporal Sequences) and three models (\texttt{LLaMA-3.1-8B}, \texttt{Mistral-7B-v0.1}, \texttt{Qwen2.5-7B}). Weighted averages (by task sample size) of correlation and accuracy are shown in Figures~\ref{fig:heatmap_base} and~\ref{fig:heatmap_instruct}. For base models, we consider two prompt templates and assign a pair of example scores $(i,j)$ per template (Figures~\ref{fig:prompt1_base}, \ref{fig:prompt2_base}). For instruction-tuned models, we assign a single example score in ${1,\ldots,10}$ and include a random baseline (Figure~\ref{fig:prompt_instruct}). We also test the instruction prompt from \citet{xiong2023can} without explicit example scores (Figure~\ref{fig:prompt_instruct_noscore}).

Two regularities emerge. First, performance and correlation tend to increase when the two exemplar scores are close, but not identical. Identical scores reduce the variance of produced linguistic confidence, which lowers instance-level association. Second, including a very large exemplar score pushes the model toward high outputs (often 8--10) and compresses variability. Balancing correlation and accuracy, the selected settings are $(5,6)$ for Base Prompt 1, $(4,5)$ for Base Prompt 2, and a single score of $2$ for the Instruct prompt.

\subsection{Ablation Studies}
\label{appendix:ablation}

We test whether eliciting linguistic confidence perturbs final-layer logits. For every model and all eight tasks, we repeat the main setup without requesting confidence, holding all other settings fixed. For base models, each prompt still contains two in-context QA exemplars. Representative prompts are shown in Figures~\ref{fig:base_withoutconf} and~\ref{fig:instruct_withoutconf}. Paired comparisons of normalized accuracy, average logits, logits variance, and ECE are reported in Table~\ref{tbl:ablation-results}.

\begin{table}[t]
\centering
\small
\setlength{\tabcolsep}{3pt}
\begin{tabular}{lrrrr}
\toprule
Metric & mean$_a$ & mean$_b$ & $t$ & $p$ \\
\midrule
\multicolumn{5}{l}{\textit{Base Prompt 2 $\rightarrow$ Base Prompt 1}} \\
Normalized Accuracy & 0.413 & 0.405 & 0.55 & 5.9e-01 \\
Average Logits & 0.596 & 0.569 & 8.22 & \underline{2.1e-13} \\
Logits Variance & 0.112 & 0.108 & 3.08 & \underline{2.5e-03} \\
ECE(logits) & 0.129 & 0.140 & -1.51 & 1.3e-01 \\
\midrule
\multicolumn{5}{l}{\textit{Base Prompt 1 $\rightarrow$ Base without Confidence}} \\
Normalized Accuracy & 0.405 & 0.405 & 0.00 & 9.9e-01 \\
Average Logits & 0.569 & 0.617 & -11.58 & \underline{1.3e-21} \\
Logits Variance & 0.108 & 0.125 & -7.76 & \underline{2.4e-12} \\
ECE(logits) & 0.140 & 0.122 & 2.96 & \underline{3.7e-03} \\
\midrule
\multicolumn{5}{l}{\textit{Base Prompt 2 $\rightarrow$ Base without Confidence}} \\
Normalized Accuracy & 0.413 & 0.405 & 0.49 & 6.2e-01 \\
Average Logits & 0.596 & 0.617 & -5.56 & \underline{1.5e-07} \\
Logits Variance & 0.112 & 0.125 & -6.29 & \underline{4.7e-09} \\
ECE(logits) & 0.129 & 0.122 & 1.18 & 2.4e-01 \\
\midrule
\multicolumn{5}{l}{\textit{Instruct Prompt $\rightarrow$ Instruct without Confidence}} \\
Normalized Accuracy & 0.367 & 0.370 & -0.17 & 8.7e-01 \\
Average Logits & 0.875 & 0.881 & -0.89 & 3.7e-01 \\
Logits Variance & 0.119 & 0.112 & 1.50 & 1.4e-01 \\
ECE(logits) & 0.286 & 0.294 & -0.57 & 5.7e-01 \\
\bottomrule
\end{tabular}
\caption{Paired t-test results comparing logits-based statistics under different prompting conditions. 
Mean values ($a,b$), $t$-statistics, and $p$-values are reported. Underlined values indicate significant differences ($p<0.05$).}
\label{tbl:ablation-results}
\end{table}

As shown in Table~\ref{tbl:ablation-results}, removing confidence elicitation leaves normalized accuracy unchanged. For base models, however, it produces small but statistically significant shifts in average logits and logits variance and, in one comparison, in logits ECE. For instruction-tuned models, none of the tested metrics changes significantly. Thus, confidence elicitation does not affect task accuracy in our experiments, but it can modestly perturb internal probability distributions for base models.

\subsection{Prompt Perturbation}
\label{appendix:prompt_perturbation}

To test sensitivity to user attitude, we prepend short cues that convey criticism, approval, or irrelevance. Each attitude uses three variants inserted before the main QA examples. We evaluate eight tasks on three diverse models (Qwen2.5-7B, Mistral-7B-v0.3, Llama-3.1-8B) and average within attitude type. We also include a calibration variant that appends ``Make sure the confidence you report reflects the softmax probabilities from your final-layer logits.'' The cue sets are listed in Figure~\ref{fig:attitude} (Appendix~\ref{appendix:prompt}). Results appear in Table~\ref{tbl:perturbation-results}.

\begin{table}[t]
\centering
\small
\setlength{\tabcolsep}{4pt}

\begin{tabular}{lrrrr}
\toprule
Metric & mean$_a$ & mean$_b$ & $t$ & $p$ \\
\midrule
\multicolumn{5}{l}{\textit{Origin $\rightarrow$ Criticism}} \\
$\mathrm{Avg}(\mathrm{Conf}(li))$ & 8.428 & 8.612 & -2.34 & \underline{0.028} \\
$r$ & 0.266 & 0.146 & 3.07 & \underline{0.0056} \\
\midrule
\multicolumn{5}{l}{\textit{Origin $\rightarrow$ Approval}} \\
$\mathrm{Avg}(\mathrm{Conf}(li))$ & 8.428 & 8.807 & -3.75 & \underline{0.0011} \\
$r$ & 0.266 & 0.130 & 2.43 & \underline{0.0235} \\
\midrule
\multicolumn{5}{l}{\textit{Origin $\rightarrow$ Irrelevant}} \\
$\mathrm{Avg}(\mathrm{Conf}(li))$ & 8.428 & 8.654 & -2.47 & \underline{0.0213} \\
$r$ & 0.266 & 0.178 & 3.14 & \underline{0.0047} \\
\midrule
\multicolumn{5}{l}{\textit{Origin $\rightarrow$ Calibration Variant}} \\
$\mathrm{Avg}(\mathrm{Conf}(li))$ & 8.428 & 8.396 & 0.27 & 0.793 \\
$r$ & 0.266 & 0.223 & 1.25 & 0.226 \\
\bottomrule
\end{tabular}
\caption{Paired t-test results comparing the effect of different attitude and calibration variants on average linguistic confidence $\mathrm{Avg}(\mathrm{Conf}(li))$ and its Pearson correlation with logits-based confidence $r$. Underlined values indicate significant differences ($p<0.05$).}
\label{tbl:perturbation-results}
\end{table}

As shown in Table~\ref{tbl:perturbation-results}, attitude cues consistently raise reported linguistic confidence. Approval has the largest effect, criticism is also positive, and even irrelevant cues increase averages. At the same time, the association with logits decreases, which indicates that inflation in reported confidence is not grounded in internal probabilities. The calibration instruction does not significantly change either average confidence or correlation. Overall, prompt tone can inflate stated confidence and weaken its association with internal probabilities.

\section{Generation Task Details}
\label{appendix:gen_details}
\subsection{Prompt Design for Generation Tasks}
\label{appendix:gen_prompt}

We follow the prompting setup of \citet{kuhn2023semantic}. To elicit linguistic confidence, we additionally append a confidence query instruction following \citet{xiong2023can}. Specifically, we include random example confidence scores ranging from 0 to 10 after each question-answer pair in CoQA or after each demonstration example in TriviaQA. The model-generated confidence is extracted and averaged across the five sampled generations for each prompt.

Prompt examples are shown in Figure~\ref{fig:coqa} and Figure~\ref{fig:triviaqa}.

\subsection{Full Results on Generation Tasks}
\label{appendix:gen_results}
We calculate the semantic entropy and the average linguistic confidence of the five generated sequences for each prompt, and then compute the Pearson correlation between them (the semantic entropy is negated when computing the correlation, since a higher entropy indicates lower model confidence). The results are presented in Table~\ref{tab:gen_results}. 
\begin{table*}[t] 
\centering \setlength{\tabcolsep}{5pt} \small 
\begin{tabular}{lcccc@{\hspace{8mm}}lcccc} 
\toprule 
\multicolumn{5}{c}{\textbf{Llama Family}} & \multicolumn{5}{c}{\textbf{Mistral / Mixtral Family}} \\ 
\cmidrule(lr){1-5}\cmidrule(lr){6-10} Model & \multicolumn{2}{c}{CoQA ($r$)} & \multicolumn{2}{c}{TriviaQA ($r$)} & Model & \multicolumn{2}{c}{CoQA ($r$)} & \multicolumn{2}{c}{TriviaQA ($r$)} \\
\cmidrule(lr){2-3}\cmidrule(lr){4-5}\cmidrule(lr){7-8}\cmidrule(lr){9-10} & Base & Instruct & Base & Instruct & & Base & Instruct & Base & Instruct \\ 
\midrule Llama-2-7B & 0.056 & \textbf{0.063} & \underline{0.211} & \textbf{\underline{0.433}} & Mistral-7B-v0.1 & $-$0.040 & \textbf{$-$0.017} & \underline{0.288} & \textbf{\underline{0.392}} \\ 
Llama-2-13B & \underline{$-$0.148} & \textbf{0.064} & \underline{0.224} & \textbf{\underline{0.469}} & Mistral-7B-v0.2 & $-$0.008 & \textbf{\underline{0.213}} & \underline{0.251} & \textbf{\underline{0.338}} \\
Llama-3.1-8B & $-$0.015 & \textbf{\underline{0.161}} & 0.004 & \textbf{\underline{0.318}} & Mistral-7B-v0.3 & $-$0.048 & \textbf{0.012} & 0.033 & \textbf{\underline{0.549}} \\ 
Llama-3.2-3B & 0.000 & \textbf{\underline{0.245}} & \underline{0.261} & \textbf{\underline{0.591}} & Mistral-Nemo-2407 & $-$0.050 & \textbf{$-$0.015} & \underline{0.355} & \textbf{\underline{0.472}} \\ 
Llama-3-8B & \textbf{0.008} & $-$0.006 & $-$0.034 & \textbf{\underline{0.153}} & Mixtral-8$\times$7B-v0.1 & $-$0.024 & \textbf{0.077} & \textbf{\underline{0.346}} & \underline{0.334} \\ 
\bottomrule 
\end{tabular} 
\caption{Pearson correlation ($r$) between semantic-entropy-based and linguistic confidence on CoQA and TriviaQA. Higher correlations within each base/instruction-tuned pair are highlighted in bold; underlined values indicate statistical significance ($p<0.05$).} 
\label{tab:gen_results} 
\vspace{-2mm} 
\end{table*}

Unlike logits-based confidence, which directly reflects the model's internal probability distribution over tokens, \textit{semantic entropy} is designed to capture uncertainty in generative tasks where the exact answer span or key token positions are difficult to pinpoint. It measures how semantically diverse the generated responses are rather than how peaked the token-level probabilities appear. To further illustrate the distinction between these two uncertainty measures, we also compare logits-based confidence and semantic entropy on several classification tasks, as detailed in Appendix~\ref{appendix:classification_entropy}.

\subsection{Semantic Entropy and Logits-Based Confidence}
\label{appendix:classification_entropy}

Unlike logits-based confidence, which directly reflects token-level probability distributions, semantic entropy captures uncertainty in generative settings where exact answer spans or salient token positions are difficult to identify. It quantifies uncertainty through semantic diversity among generated responses rather than probability concentration over tokens.

To further illustrate the distinction between these uncertainty measures, we additionally compare logits-based confidence and semantic entropy on several classification tasks.

For an $N$-option classification task, we first obtain the softmax probabilities from the model's final-layer logits, denoted as $p_1, p_2, \dots, p_N$.
The semantic entropy is then computed as
\begin{equation}
SE = -\sum_{i=1}^{N} p_i \log(p_i).
\end{equation}
The corresponding linguistic confidence for entropy is the average score obtained by inserting each candidate option into the prompt and asking the model for its confidence individually.
We use \texttt{Qwen2.5-14B} to compute correlations between linguistic confidence and (1) logits-based confidence and (2) negated semantic entropy under two prompt templates. Results are shown in Table~\ref{tab:entropy_corr}.

\begin{table*}[t]
\centering
\setlength{\tabcolsep}{5pt} 
\begin{tabular}{lrrrr}
\toprule
\textbf{Task} &
\multicolumn{2}{c}{\textbf{Base Prompt 1}} &
\multicolumn{2}{c}{\textbf{Base Prompt 2}} \\
\cmidrule(lr){2-3}\cmidrule(lr){4-5}
 & \textbf{Logits Corr.} & \textbf{Entropy Corr.} &
   \textbf{Logits Corr.} & \textbf{Entropy Corr.} \\
\midrule
Conceptual Combinations & \textbf{0.4441} & -0.3111 & \textbf{0.8189 } & -0.1970 \\
Ruin Names              & \textbf{0.4479} & -0.1096 & \textbf{0.4142 } & -0.3364 \\
Temporal Sequences      & \textbf{0.1246} &  0.0103 & \textbf{0.0431 } & -0.0401 \\
CoLA                    & \textbf{0.4469} &  0.2283 & \textbf{0.5672 } &  0.1077 \\
Cause and Effect        & \textbf{0.1413} &  0.0133 & \textbf{-0.0853}  & -0.3545 \\
MMLU                    & \textbf{0.3909} & -0.1662 & \textbf{0.3345 } & -0.1089 \\
QNLI                    & \textbf{0.1786} & -0.0469 & \textbf{0.1142 } &  0.0389 \\
QQP                     & \textbf{0.5777} & -0.3634 & \textbf{0.6227 } & -0.1375 \\
\bottomrule
\end{tabular}
\caption{Correlation between linguistic confidence and internal metrics (logits-based confidence vs. negated semantic entropy) on \texttt{Qwen2.5-14B} under two base prompt templates.}
\label{tab:entropy_corr}
\end{table*}
Across the 16 task-prompt settings, logits-based confidence is positively correlated with linguistic confidence in 15, whereas the association with negated option-level predictive entropy is generally weaker and often negative. This suggests that token-probability confidence and entropy capture different internal views of reliability. Linguistic confidence is closer to the former in these classification settings.

\section{Regression Model} \label{app:regression}

 \paragraph{Outcomes and Predictors.}
 We independently model three continuous outcomes:
\textsc{Dist\_linguistic-logits},
\textsc{ECE\_linguistic-logits}, and
\textsc{Correlation}.

Predictors include model metadata
(Model\_family, Model\_version,
Model\_type, Model\_size),
task characteristics (Task, Num\_choices),
distributional statistics of linguistic and logits confidence
(Linguistic\_mean,
Linguistic\_std,
Logits\_mean,
Logits\_std),
and performance-related measures
(Accuracy, Norm\_acc,
ECE\_linguistic, ECE\_logits).

Categorical fields, including Model\_family, Model\_version, Model\_type and Task, are one-hot encoded with a dropped reference. Zero-variance columns are removed and an intercept is included.
 
 The full OLS coefficients for the three outcomes are shown in Tables~\ref{tab:ols-dist}, \ref{tab:ols-ece}, and \ref{tab:ols-corr}. We also report ridge standardized effects for each outcome in Tables~\ref{tab:ridge-corr}, \ref{tab:ridge-dist}, and \ref{tab:ridge-ececl} to rank effect sizes under collinearity.
 
 \begin{table*}[p]
 \centering
 \scriptsize
 \setlength{\tabcolsep}{3pt} 
 \resizebox{\textwidth}{!}{%
 \begin{tblr}{
   colspec = {l r r r r r r r},
   hline{1,Z} = {1pt},
   hline{2}   = {0.5pt},
 }
 Predictor & Coef & Std.\(\beta\) & SE & \(t\) & \(p\) & CI\(_{2.5}\) & CI\(_{97.5}\) \\
 Linguistic\_std & 0.080817 & 0.780064 & 0.005486 & 14.733 & 3.02e-38 & 0.070026 & 0.091607 \\
 Num\_choices & 0.029610 & 0.286977 & 0.005590 & 5.297 & 2.14e-07 & 0.018613 & 0.040606 \\
 Logits\_mean & -0.281884 & -0.558972 & 0.053266 & -5.292 & 2.19e-07 & -0.386661 & -0.177108 \\
 Linguistic\_mean & 0.019635 & 0.227861 & 0.005047 & 3.890 & 1.21e-04 & 0.009707 & 0.029563 \\
 Task\_QNLI & 0.041021 & 0.131019 & 0.015235 & 2.693 & 0.0074 & 0.011053 & 0.070989 \\
 Model\_version\_3 & 0.031787 & 0.127700 & 0.015214 & 2.089 & 0.0374 & 0.001860 & 0.061713 \\
 Norm\_acc & 0.159594 & 0.434841 & 0.080033 & 1.994 & 0.0469 & 0.002165 & 0.317022 \\
 Task\_MMLU & -0.035128 & -0.113260 & 0.017947 & -1.957 & 0.0511 & -0.070430 & 0.000173 \\
 Model\_version\_2 & 0.029361 & 0.091958 & 0.015618 & 1.880 & 0.0610 & -0.001361 & 0.060083 \\
 Task\_conceptual\_combinations & -0.031565 & -0.100818 & 0.018539 & -1.703 & 0.0896 & -0.068033 & 0.004902 \\
 Model\_family\_mistral & 0.025996 & 0.118525 & 0.015467 & 1.681 & 0.0937 & -0.004427 & 0.056420 \\
 Task\_QQP & 0.024429 & 0.078764 & 0.015129 & 1.615 & 0.1073 & -0.005330 & 0.054188 \\
 Accuracy & -0.203839 & -0.373949 & 0.127119 & -1.604 & 0.1098 & -0.453889 & 0.046210 \\
 ECE\_linguistic & 0.066674 & 0.095053 & 0.044742 & 1.490 & 0.1371 & -0.021336 & 0.154685 \\
 Task\_temporal\_sequences & -0.027931 & -0.089210 & 0.019908 & -1.403 & 0.1615 & -0.067090 & 0.011229 \\
 Model\_version\_3.2 & 0.024137 & 0.058129 & 0.017625 & 1.369 & 0.1718 & -0.010532 & 0.058805 \\
 Task\_ruin\_names & -0.019467 & -0.062176 & 0.021312 & -0.913 & 0.3617 & -0.061389 & 0.022455 \\
 Model\_version\_3.1 & 0.013335 & 0.032116 & 0.018014 & 0.740 & 0.4597 & -0.022100 & 0.048770 \\
 Task\_cause\_and\_effect & 0.010221 & 0.032647 & 0.015239 & 0.671 & 0.5028 & -0.019754 & 0.040196 \\
 Model\_version\_0.3 & 0.009882 & 0.023799 & 0.018287 & 0.540 & 0.5893 & -0.026089 & 0.045853 \\
 Model\_version\_nemo & -0.009373 & -0.022573 & 0.018757 & -0.500 & 0.6176 & -0.046268 & 0.027522 \\
 Model\_version\_0.2 & 0.007431 & 0.017895 & 0.018672 & 0.398 & 0.6909 & -0.029298 & 0.044159 \\
 Model\_family\_qwen & 0.007809 & 0.036160 & 0.022527 & 0.347 & 0.7291 & -0.036503 & 0.052121 \\
 Logits\_std & 0.025886 & 0.015045 & 0.085453 & 0.303 & 0.7621 & -0.142205 & 0.193976 \\
 Model\_type\_instruct & 0.005372 & 0.023738 & 0.021440 & 0.251 & 0.8023 & -0.036802 & 0.047546 \\
 Model\_version\_2.5 & 0.004281 & 0.016544 & 0.019168 & 0.223 & 0.8234 & -0.033423 & 0.041985 \\
 Model\_size & -0.000230 & -0.007926 & 0.001373 & -0.168 & 0.8669 & -0.002931 & 0.002471 \\
 ECE\_logits & 0.004661 & 0.006237 & 0.041906 & 0.111 & 0.9115 & -0.077770 & 0.087091 \\
 const & 0.128896 &  & 0.064189 & 2.008 & 0.0454 & 0.002633 & 0.255160 \\
 \end{tblr}%
 } 
 \caption{Full OLS coefficients for \textsc{Dist\_linguistic-logits} under specified predictors.}
 \label{tab:ols-dist}
 \end{table*}

 \begin{table*}[p]
 \centering
 \scriptsize
 \setlength{\tabcolsep}{3pt} 
 \resizebox{\textwidth}{!}{%
 \begin{tblr}{
   colspec = {l r r r r r r r},
   hline{1,Z} = {1pt},
   hline{2}   = {0.5pt},
 }
 Predictor & Coef & Std.\(\beta\) & SE & \(t\) & \(p\) & CI\(_{2.5}\) & CI\(_{97.5}\) \\
 Logits\_mean & -0.457089 & -0.911912 & 0.045934 & -9.951 & 1.29e-20 & -0.547444 & -0.366734 \\
 Linguistic\_mean & 0.033871 & 0.395460 & 0.004352 & 7.782 & 8.89e-14 & 0.025309 & 0.042432 \\
 Linguistic\_std & 0.032748 & 0.318015 & 0.004731 & 6.923 & 2.26e-11 & 0.023443 & 0.042053 \\
 Logits\_std & -0.365030 & -0.213445 & 0.073691 & -4.954 & 1.16e-06 & -0.509985 & -0.220076 \\
 Num\_choices & 0.021411 & 0.208775 & 0.004821 & 4.441 & 1.22e-05 & 0.011928 & 0.030894 \\
 ECE\_linguistic & 0.159993 & 0.229478 & 0.038584 & 4.147 & 4.28e-05 & 0.084096 & 0.235890 \\
 Model\_family\_mistral & 0.034745 & 0.159377 & 0.013338 & 2.605 & 0.0096 & 0.008509 & 0.060981 \\
 Model\_version\_2 & 0.030157 & 0.095024 & 0.013468 & 2.239 & 0.0258 & 0.003664 & 0.056650 \\
 Task\_MMLU & -0.033979 & -0.110219 & 0.015476 & -2.196 & 0.0288 & -0.064422 & -0.003536 \\
 Task\_temporal\_sequences & -0.035044 & -0.112611 & 0.017168 & -2.041 & 0.0420 & -0.068814 & -0.001275 \\
 Model\_version\_nemo & -0.032071 & -0.077707 & 0.016175 & -1.983 & 0.0482 & -0.063888 & -0.000254 \\
 Model\_size & 0.002269 & 0.078583 & 0.001184 & 1.917 & 0.0561 & -0.000060 & 0.004598 \\
 Model\_version\_3.2 & 0.029128 & 0.070575 & 0.015199 & 1.916 & 0.0562 & -0.000769 & 0.059024 \\
 ECE\_logits & -0.068931 & -0.092804 & 0.036138 & -1.907 & 0.0573 & -0.140016 & 0.002154 \\
 Norm\_acc & 0.119477 & 0.327515 & 0.069017 & 1.731 & 0.0843 & -0.016283 & 0.255237 \\
 Task\_QNLI & 0.022519 & 0.072360 & 0.013138 & 1.714 & 0.0875 & -0.003325 & 0.048362 \\
 Model\_version\_3 & 0.020161 & 0.081486 & 0.013120 & 1.537 & 0.1253 & -0.005647 & 0.045968 \\
 Task\_conceptual\_combinations & -0.023002 & -0.073915 & 0.015988 & -1.439 & 0.1512 & -0.054450 & 0.008446 \\
 Model\_version\_0.2 & -0.021393 & -0.051835 & 0.016102 & -1.329 & 0.1849 & -0.053067 & 0.010280 \\
 Model\_family\_qwen & 0.025702 & 0.119743 & 0.019426 & 1.323 & 0.1867 & -0.012511 & 0.063915 \\
 Accuracy & -0.114829 & -0.211939 & 0.109622 & -1.048 & 0.2956 & -0.330462 & 0.100803 \\
 Model\_version\_0.3 & -0.016109 & -0.039032 & 0.015770 & -1.022 & 0.3078 & -0.047129 & 0.014911 \\
 Model\_version\_3.1 & 0.014661 & 0.035523 & 0.015535 & 0.944 & 0.3460 & -0.015897 & 0.045219 \\
 Task\_ruin\_names & -0.016377 & -0.052627 & 0.018379 & -0.891 & 0.3735 & -0.052529 & 0.019774 \\
 Task\_QQP & 0.010294 & 0.033392 & 0.013046 & 0.789 & 0.4306 & -0.015369 & 0.035957 \\
 Model\_version\_2.5 & -0.009743 & -0.037882 & 0.016530 & -0.589 & 0.5560 & -0.042257 & 0.022772 \\
 Task\_cause\_and\_effect & 0.006342 & 0.020379 & 0.013141 & 0.483 & 0.6297 & -0.019507 & 0.032191 \\
 Model\_type\_instruct & 0.004896 & 0.021766 & 0.018489 & 0.265 & 0.7913 & -0.031473 & 0.041265 \\
 const & 0.119108 &  & 0.055354 & 2.152 & 0.0321 & 0.010224 & 0.227992 \\
 \end{tblr}%
 } 
 \caption{Full OLS coefficients for \textsc{ECE\_linguistic-logits} under specified predictors.}
 \label{tab:ols-ece}
 \end{table*}

 \begin{table*}[p]
 \centering
 \scriptsize
 \setlength{\tabcolsep}{3pt} 
 \resizebox{\textwidth}{!}{%
 \begin{tblr}{
   colspec = {l r r r r r r r},
   hline{1,Z} = {1pt},
   hline{2}   = {0.5pt},
 }
 Predictor & Coef & Std.\(\beta\) & SE & \(t\) & \(p\) & CI\(_{2.5}\) & CI\(_{97.5}\) \\
 Task\_QNLI & -0.155245 & -0.251365 & 0.038621 & -4.020 & 7.27e-05 & -0.231227 & -0.079263 \\
 Model\_size & 0.011347 & 0.194214 & 0.003532 & 3.213 & 0.0014 & 0.004398 & 0.018296 \\
 Linguistic\_std & 0.041036 & 0.194311 & 0.014786 & 2.775 & 0.0058 & 0.011945 & 0.070126 \\
 Task\_cause\_and\_effect & -0.107774 & -0.171142 & 0.039022 & -2.762 & 0.0061 & -0.184544 & -0.031003 \\
 Task\_QQP & -0.102290 & -0.167177 & 0.038298 & -2.671 & 0.0080 & -0.177638 & -0.026943 \\
 Linguistic\_mean & 0.028679 & 0.164087 & 0.013086 & 2.192 & 0.0291 & 0.002934 & 0.054423 \\
 Model\_family\_qwen & 0.124865 & 0.289072 & 0.059373 & 2.103 & 0.0362 & 0.008055 & 0.241674 \\
 Task\_temporal\_sequences & -0.093571 & -0.145567 & 0.052304 & -1.789 & 0.0746 & -0.196474 & 0.009332 \\
 Norm\_acc & 0.349829 & 0.466761 & 0.209766 & 1.668 & 0.0963 & -0.062861 & 0.762519 \\
 Accuracy & -0.541172 & -0.476354 & 0.335281 & -1.614 & 0.1075 & -1.200797 & 0.118454 \\
 Task\_MMLU & -0.074955 & -0.121363 & 0.046563 & -1.610 & 0.1084 & -0.166562 & 0.016653 \\
 ECE\_linguistic & -0.176518 & -0.115766 & 0.115576 & -1.527 & 0.1277 & -0.403899 & 0.050864 \\
 Model\_version\_nemo & -0.072938 & -0.083811 & 0.049779 & -1.465 & 0.1438 & -0.170872 & 0.024995 \\
 Model\_family\_mistral & 0.054200 & 0.122749 & 0.039799 & 1.362 & 0.1742 & -0.024100 & 0.132500 \\
 Logits\_mean & -0.183698 & -0.182976 & 0.139718 & -1.315 & 0.1895 & -0.458576 & 0.091181 \\
 ECE\_logits & -0.142873 & -0.091102 & 0.109132 & -1.309 & 0.1914 & -0.357577 & 0.071832 \\
 Model\_version\_3.1 & 0.055296 & 0.067613 & 0.046194 & 1.197 & 0.2322 & -0.035586 & 0.146178 \\
 Model\_type\_instruct & 0.063561 & 0.139207 & 0.055246 & 1.151 & 0.2508 & -0.045129 & 0.172252 \\
 Model\_version\_3.2 & 0.049355 & 0.057957 & 0.045985 & 1.073 & 0.2840 & -0.041116 & 0.139825 \\
 Task\_conceptual\_combinations & 0.033876 & 0.052701 & 0.048393 & 0.700 & 0.4844 & -0.061331 & 0.129084 \\
 Model\_version\_0.3 & 0.026995 & 0.032363 & 0.047290 & 0.571 & 0.5685 & -0.066042 & 0.120032 \\
 Logits\_std & 0.112563 & 0.032396 & 0.227371 & 0.495 & 0.6209 & -0.334763 & 0.559889 \\
 Model\_version\_3 & -0.015929 & -0.031757 & 0.039442 & -0.404 & 0.6866 & -0.093527 & 0.061669 \\
 Num\_choices & 0.004230 & 0.020402 & 0.014322 & 0.295 & 0.7679 & -0.023946 & 0.032407 \\
 Model\_version\_2.5 & -0.013040 & -0.025490 & 0.049900 & -0.261 & 0.7940 & -0.111212 & 0.085131 \\
 Task\_ruin\_names & 0.013881 & 0.021594 & 0.055952 & 0.248 & 0.8042 & -0.096199 & 0.123960 \\
 Model\_version\_2 & -0.006998 & -0.010770 & 0.041111 & -0.170 & 0.8649 & -0.087878 & 0.073883 \\
 Model\_version\_0.2 & 0.004411 & 0.005393 & 0.047652 & 0.093 & 0.9263 & -0.089340 & 0.098161 \\
 const & 0.122884 &  & 0.167205 & 0.735 & 0.4629 & -0.206073 & 0.451841 \\
 \end{tblr}%
 } 
 \caption{Full OLS coefficients for \textsc{Correlation} under specified predictors.}
 \label{tab:ols-corr}
 \end{table*}

 \begin{table}[t]
 \centering
 \small
 \renewcommand{\arraystretch}{1.1}
 \begin{tabular}{l r}
 \toprule
 \textbf{Predictor} & \textbf{Std.($\beta$)} \\
 \midrule
 Model\_size & -0.012746 \\
 Num\_choices & 0.143658 \\
 Linguistic\_mean & 0.201203 \\
 Linguistic\_std & 0.741303 \\
 Logits\_mean & -0.496499 \\
 Logits\_std & 0.009902 \\
 ECE\_logits & -0.006060 \\
 Norm\_acc & 0.227175 \\
 ECE\_linguistic & 0.123461 \\
 Accuracy & -0.158557 \\
 Model\_family\_mistral & 0.017754 \\
 Model\_family\_qwen & 0.014370 \\
 Model\_version\_0.2 & 0.013036 \\
 Model\_version\_0.3 & 0.020893 \\
 Model\_version\_2 & 0.015764 \\
 Model\_version\_2.5 & -0.057565 \\
 Model\_version\_3 & 0.037234 \\
 Model\_version\_3.1 & -0.021622 \\
 Model\_version\_3.2 & -0.002226 \\
 Model\_version\_nemo & -0.030178 \\
 Model\_type\_instruct & 0.011662 \\
 Task\_MMLU & 0.034117 \\
 Task\_QNLI & 0.121791 \\
 Task\_QQP & 0.063017 \\
 Task\_cause\_and\_effect & 0.023452 \\
 Task\_conceptual\_combinations & 0.043844 \\
 Task\_ruin\_names & 0.086675 \\
 Task\_temporal\_sequences & 0.053004 \\
 \bottomrule
 \end{tabular}
 \caption{Full ridge (CV) standardized coefficients for \textsc{Dist\_linguistic-logits}.}
 \label{tab:ridge-dist}
 \end{table}
 
 \begin{table}[t]
 \centering
 \small
 \begin{tabular}{l r}
 \toprule
 \textbf{Predictor} & \textbf{Std.($\beta$)} \\
 \midrule
 Model\_size & 0.071043 \\
 Num\_choices & 0.080202 \\
 Linguistic\_mean & 0.381558 \\
 Linguistic\_std & 0.302399 \\
 Logits\_mean & -0.843727 \\
 Logits\_std & -0.215614 \\
 ECE\_logits & -0.099016 \\
 Norm\_acc & 0.232477 \\
 ECE\_linguistic & 0.240305 \\
 Accuracy & -0.118653 \\
 Model\_family\_mistral & 0.063824 \\
 Model\_family\_qwen & 0.083243 \\
 Model\_version\_0.2 & -0.055676 \\
 Model\_version\_0.3 & -0.041310 \\
 Model\_version\_2 & 0.028802 \\
 Model\_version\_2.5 & -0.102369 \\
 Model\_version\_3 & 0.001939 \\
 Model\_version\_3.1 & -0.013762 \\
 Model\_version\_3.2 & 0.016518 \\
 Model\_version\_nemo & -0.077676 \\
 Model\_type\_instruct & -0.011264 \\
 Task\_MMLU & 0.004494 \\
 Task\_QNLI & 0.068909 \\
 Task\_QQP & 0.028014 \\
 Task\_cause\_and\_effect & 0.017697 \\
 Task\_conceptual\_combinations & 0.038977 \\
 Task\_ruin\_names & 0.069560 \\
 Task\_temporal\_sequences & 0.008612 \\
 \bottomrule
 \end{tabular}
 \caption{Full ridge (CV) standardized coefficients for \textsc{ECE\_linguistic-logits}.}
 \label{tab:ridge-ececl}
 \end{table}
 
 \begin{table}[t]
 \centering
 \small
 \begin{tabular}{l r}
 \toprule
 \textbf{Predictor} & \textbf{Std.($\beta$)} \\
 \midrule
 Model\_size & 0.113674 \\
 Num\_choices & 0.109605 \\
 Linguistic\_mean & 0.092039 \\
 Linguistic\_std & 0.087174 \\
 Logits\_mean & -0.007872 \\
 Logits\_std & 0.036264 \\
 ECE\_logits & -0.071479 \\
 Norm\_acc & 0.075122 \\
 ECE\_linguistic & -0.038211 \\
 Accuracy & -0.030490 \\
 Model\_family\_mistral & 0.029550 \\
 Model\_family\_qwen & 0.079445 \\
 Model\_version\_0.2 & -0.003542 \\
 Model\_version\_0.3 & 0.023032 \\
 Model\_version\_2 & -0.052616 \\
 Model\_version\_2.5 & 0.030971 \\
 Model\_version\_3 & -0.029987 \\
 Model\_version\_3.1 & 0.023491 \\
 Model\_version\_3.2 & -0.013133 \\
 Model\_version\_nemo & -0.061904 \\
 Model\_type\_instruct & 0.056131 \\
 Task\_MMLU & -0.009412 \\
 Task\_QNLI & -0.155651 \\
 Task\_QQP & -0.099535 \\
 Task\_cause\_and\_effect & -0.099387 \\
 Task\_conceptual\_combinations & 0.114804 \\
 Task\_ruin\_names & 0.112571 \\
 Task\_temporal\_sequences & -0.047692 \\
 \bottomrule
 \end{tabular}
 \caption{Full ridge (CV) standardized coefficients for \textsc{Correlation}.}
 \label{tab:ridge-corr}
 \end{table}
\end{document}